\PassOptionsToPackage{table}{xcolor}
\documentclass[]{youtu}

\usepackage{natbib}

\usepackage{times}
\usepackage{latexsym}
\usepackage[T1]{fontenc}
\usepackage[utf8]{inputenc}
\usepackage{bbold}
\usepackage{inconsolata}

\usepackage{microtype}
\usepackage{parskip}
\usepackage{ragged2e}
\usepackage{amsmath}
\usepackage{amssymb}
\usepackage{amsfonts}
\usepackage{nicefrac}
\usepackage{pifont}

\usepackage{xcolor}
\definecolor{softpurple}{RGB}{120,80,160}
\definecolor{rapobg}{RGB}{230,241,249}
\definecolor{promptboxbg}{RGB}{248,250,252}
\definecolor{promptboxframe}{RGB}{56,84,145}

\usepackage{graphicx}
\usepackage{float}
\usepackage{adjustbox}
\usepackage{wrapfig}

\usepackage{booktabs}
\usepackage{array}
\usepackage{multirow}
\usepackage{tabularx}
\usepackage{makecell}
\usepackage{arydshln}
\newcolumntype{Y}{>{\raggedright\arraybackslash}p{0.24\textwidth}}

\usepackage{caption}
\usepackage{subcaption}
\usepackage{enumitem}
\usepackage{fvextra}
\usepackage{listings}
\usepackage{xurl}

\lstdefinestyle{promptstyle}{
  basicstyle=\ttfamily\scriptsize,
  breaklines=true,
  breakatwhitespace=false,
  columns=fullflexible,
  keepspaces=true,
  showstringspaces=false,
  frame=none,
}

\newtcolorbox{promptbox}[1][]{
  breakable,
  colback=promptboxbg,
  colframe=promptboxframe,
  coltitle=white,
  fonttitle=\bfseries,
  title={#1},
  arc=3mm,
  boxrule=0.5pt,
  left=2mm, right=2mm, top=2mm, bottom=2mm,
  fontupper=\scriptsize\ttfamily,
}

\usepackage[ruled,vlined]{algorithm2e}
\usepackage{algorithmic}

\title{Improving General Role-Playing Agents via Psychology-Grounded Reasoning and Role-Aware Policy Optimization}
\author{\small Zhenhua Xu\textsuperscript{1,*}, Dongsheng Chen\textsuperscript{2,*}, Jian Li\textsuperscript{2}, Yitong Lin\textsuperscript{1}, Zhebo Wang\textsuperscript{1}, Jiafu Wu\textsuperscript{2}, Yizhang Jin\textsuperscript{2}, \mbox{Chengjie Wang\textsuperscript{2}}, Meng Han\textsuperscript{1,3,$\dagger$}, Yabiao Wang\textsuperscript{1,2,$\ddagger$}}
\affiliation{$^{1}$Zhejiang University, $^{2}$Tencent Youtu Lab, $^{3}$GenTel.io}

\date{2026}
\correspondence{mhan@zju.edu.cn}
\projectleader{caseywang@tencent.com}

\youtufinalcopy

\begin{document}
\vspace{0.6cm}
\abstract{
Building general-purpose role-playing agents that faithfully portray any character from a natural-language profile remains challenging.
  Explicit reasoning can help models deliberate before responding, but generic reasoning does not necessarily preserve character fidelity.
  We therefore propose \textbf{Psy-CoT}, a psychology-grounded chain-of-thought framework that decomposes pre-response reasoning into three role-specific steps---\emph{Interaction Perception}, \emph{Psychological Empathy}, and \emph{Logical Construction}---so that the model \emph{thinks dynamically} from the profile rather than merely mimicking surface patterns.
  While structured reasoning provides a foundation, it alone is insufficient; reinforcement learning is essential to further align the model with character fidelity. However, we observe that under LLM-based reward models, both generic phrases that hack the reward model and genuinely role-specific phrases receive identical gradient signals---this hacking accumulates over training, misleading the model into treating both as equally optimal choices.
  To address this, we propose \textbf{Role-Aware Policy Optimization (RAPO)}, which uses profile-induced entropy differences to weight gradients asymmetrically---amplifying role-specific tokens under positive advantage while attenuating them under negative advantage.
  Experiments on CoSER, CharacterBench, and CharacterEval demonstrate that Psy-CoT outperforms existing role-playing CoT methods, and RAPO consistently surpasses GRPO across multiple model scales.
}

\maketitle
\fancyhead[R]{\tencentsanswseven{\texttt{RAPO}}}
\begingroup\renewcommand\thefootnote{}\footnotetext{* Equal contribution. $\dagger$ Corresponding author. $\ddagger$ Project Leader.\\ Email: \texttt{\{xuzhenhua0326,mhan\}@zju.edu.cn}}\endgroup

\section{Introduction}
\label{sec:introduction}

The strong reasoning and generation capabilities of large language models (LLMs)~\citep{li2026improvingsearchagentline, novikov2025alphaevolve, jiao2026policythoughtsscalingllm, wang2026icpoillocutioncalibratedpolicyoptimization} make them natural candidates for \emph{role-playing}---adopting and sustaining a specific persona throughout an interaction.
A particularly compelling and practical variant is the \emph{general-purpose role-playing agent} (RPA): a single model that portrays \emph{any} character given only a natural-language role profile at inference time, underpinning applications such as virtual companionship, interactive storytelling, and game NPCs~\citep{chenOscarsAITheater2025,tsengTwoTalesPersona2024,chenDesignGuidelineRPA2025,cheng2025psymemfinegrainedpsychologicalalignment,xiao2026dreamllmbaseddynamicroleplaying,li2026memrolememoriesdecisionslonghorizon}.

Explicit reasoning has improved LLM performance on a range of complex tasks~\citep{wei2022chain,shao2024deepseekmath}, making it a natural candidate for enhancing role-playing agents.
However, these gains do not automatically transfer to role-playing: \citet{fengReasoningDoesNot2025,tangThinkingCharacterAdvancing2025} report that directly applying generic reasoning can impair role-playing performance, identifying attention diversion and linguistic style drift as key challenges.
Faithful portrayal requires reasoning from the character's perspective, including their subjective beliefs, emotions, and sometimes irrational choices.
Recent work has therefore explored role-specific reasoning.
CB-CoT~\citep{liuCogDualEnhancingDual2025}, for example, organizes reasoning into structured fields describing the situation and the character's internal state.
These fields help the model consider important aspects of the character and scene.
Our focus is on how these aspects work together to shape a response: how the character's understanding of an interaction gives rise to a psychological reaction, and how that reaction guides what the character chooses to say or do.
We therefore organize reasoning as a connected process, in which each step builds on the previous one before the response is generated.

To this end, we propose \textbf{Psy-CoT}, a psychology-grounded chain-of-thought framework that organizes pre-response reasoning into three linked stages: \emph{Interaction Perception}, \emph{Psychological Empathy}, and \emph{Logical Construction}.
Interaction Perception, grounded in Theory of Mind~\citep{chen-etal-2025-theory}, first guides the model to understand the situation and infer the interlocutor's underlying intentions and mental state.
Psychological Empathy, informed by cognitive appraisal theory~\citep{lazarus1991emotion} and emotion regulation theory~\citep{gross1998emerging}, then uses this understanding to reason about how the character's background and flaws shape their emotional reaction and how to express, hide, or use that emotion.
Logical Construction, inspired by goal-directed behavior theory~\citep{locke2002building}, finally builds on this psychological reaction to select a short-term objective and plan what the character would \emph{naturally} say or do---even if impulsive or suboptimal.

Psy-CoT provides a structured process for pre-response reasoning.
To help the model better follow this process and use it to guide its responses, we further apply reinforcement learning with feedback on reasoning quality and character fidelity.
However, higher rewards do not always translate into stronger character fidelity.
During training, we repeatedly observe that agents continue to generate generic, templated actions, inner thoughts, and dialogue that could fit many different personas, even as their rewards improve.
Such expressions may receive high scores from an LLM-based reward model while remaining weakly grounded in the portrayed character's beliefs, emotions, or motivations.

Existing RL methods for role-playing largely focus on reward design~\citep{liuCogDualEnhancingDual2025,yeCPOAddressingReward2025,tangCharacterR1EnhancingRoleAware2026}, but their policy objectives do not explicitly distinguish tokens according to their relevance to the role profile.
Instead, a response-level advantage guides the optimization of all generated tokens.
When a response receives a positive advantage, both its character-specific choices and its generic expressions are reinforced, even if they contribute differently to character fidelity.
Conversely, a negative advantage can penalize character-specific choices together with the rest of an unsuccessful response, potentially discouraging useful exploration.
This motivates a more targeted optimization strategy: reinforce role-specific tokens more strongly in successful responses, while penalizing them less in unsuccessful responses to preserve exploration.

To address this, we propose \textbf{Role-Aware Policy Optimization (RAPO)}, which quantifies how much each token is influenced by the role profile and adjusts gradient contributions accordingly.
Concretely, before computing gradients, for each generated token $y_t$, RAPO computes the profile-induced entropy difference $s_t = H(p_t^{-}) - H(p_t^{+})$ by comparing the model's entropy with and without the profile $P$.
The magnitude $|s_t|$ measures how actively the profile shapes the token: $|s_t| \gg 0$ signals a role-specific token (either \textbf{exploiting} established character patterns when $s_t > 0$, or \textbf{exploring} role-specific behaviors when $s_t < 0$), while $|s_t| \approx 0$ marks a profile-agnostic token.
RAPO then applies \emph{asymmetric} per-token weighting: when the response receives a positive advantage, gradients on role-specific tokens are amplified so the model extracts more signal from its character-specific choices; when the advantage is negative, gradients on those same tokens are attenuated to reduce the cost of failed character-specific exploiting or exploration.
The result is a policy that better perceives the role profile during token generation and maintains a balance between exploration and exploitation of character-specific behaviors.

We validate our approach through extensive experiments.  Applying Psy-CoT to Llama-3.3-70B-Instruct yields average improvements of 5.2\% and 4.2\% on CoSER and CharacterBench, respectively, outperforming existing role-playing CoT methods.  Furthermore, training with RAPO on Qwen2.5-7B-Instruct, Qwen3-4B-Instruct, and Llama-3.1-8B-Instruct (all equipped with Psy-CoT) yields improvements of 13.7\%, 15.6\%, and 40.1\% over the untrained baseline on CoSER, with effectiveness generalizing to CharacterBench and CharacterEval, demonstrating that RAPO's profile-aware token weighting enables the model to generate more character-faithful responses.

\noindent Overall, \textbf{the main contributions of the work are as follows}:~(1)~We propose \textbf{Psy-CoT}, a psychology-grounded chain-of-thought framework that decomposes pre-response reasoning into Interaction Perception, Psychological Empathy, and Logical Construction, enabling role-specific deliberation rather than generic rationality.
(2)~We propose \textbf{RAPO}, a role-aware policy optimizer that leverages profile-induced entropy differences to weight gradients asymmetrically, amplifying role-specific signals while maintaining exploration.
(3)~Extensive experiments on three benchmarks and multiple model architectures validate the effectiveness of both Psy-CoT and RAPO.

\section{Related Work}
\label{sec:related_work}

\paragraph{Improving RPAs via Data Construction}
General role-playing demands faithful portrayal of arbitrary, user-defined or fictional characters. A prevalent approach addresses this by constructing large-scale multi-character corpora that maximize persona coverage and mitigate overfitting to fixed rosters. Representative works include ChatHaruhi~\citep{li2024chatharuhi}, DITTO~\citep{luLargeLanguageModels2024}, CharacterGLM~\citep{chen2024characterglm}, ROLEPERSONALITY~\citep{ran-etal-2024-capturing}, Crab~\citep{heCrabNovelConfigurable2025}, TAILORGEN~\citep{gaoTailorRPARetrievalBasedFramework2025}, CoSER~\citep{wang2025coser}, and AdaMARP~\citep{xu2026adamarpadaptivemultiagentinteraction}, each contributing distinct data-collection or synthesis strategies to broaden character coverage.

\paragraph{Reasoning and Reinforcement Learning for RPAs}
Recent works have begun to equip role-playing agents with explicit reasoning capabilities and refine them via RL. CogDual~\citep{liuCogDualEnhancingDual2025} introduces CB-CoT, which augments the thinking phase with \emph{situational awareness} and \emph{self-awareness} to better ground the character's internal state. COP~\citep{jiEnhancingPersonaConsistency2025} lets the model first perform self-questioning and self-answering, while Character-R1~\citep{tangCharacterR1EnhancingRoleAware2026} requires the model to analyze which aspects deserve focus and the corresponding attributes in the think phase. On the optimization side, CPO~\citep{yeCPOAddressingReward2025} proposes implicit comparisons to guide policy learning, while RAR~\citep{tangThinkingCharacterAdvancing2025} and PCL~\citep{jiEnhancingPersonaConsistency2025} promote preference learning through contrastive objectives.


\section{Design of RAPO}
\label{sec:design}

\subsection{Problem Formulation}
\label{sec:problem_formulation}

Following CoSER~\citep{wang2025coser}, HER~\citep{duHERHumanlikeReasoning2026}, and AdaMARP~\citep{xu2026adamarpadaptivemultiagentinteraction}, we view role-playing as an open-ended, multi-turn interaction in which character models, each instantiated from a written role specification, converse with a user and with each other to jointly unfold a story.  An episode begins with an initial role set $\mathcal{R}=\{r_1,\dots,r_N\}$ (the user treated as a distinguished member whose profile may be partially unspecified), each role $r_i$ equipped with a natural-language \emph{role profile} $P_i$ (full schema in Appendix~\ref{app:role_profile}).  Together with an initial scene, these profiles form the initial state $s_0$.  An external orchestrator selects the next speaker from $\mathcal{R}$ at every turn; in single-role settings, this degenerates to deterministic user--character alternation.  The goal is to bring the written character to life across arbitrarily long interactions, requiring simultaneous \textit{role consistency}, \textit{interaction competence}, and \textit{narrative advancement}. Each non-user role is instantiated by a shared policy $\pi_\theta$ conditioned on its profile and the shared history.  Following CoSER~\citep{wang2025coser} and HER~\citep{duHERHumanlikeReasoning2026}, a role's per-turn response interleaves \textbf{Thought} (inner monologue, wrapped in \texttt{[...]}), \textbf{Action} (externally observable cues such as gestures and expressions, wrapped in \texttt{(...)}), and \textbf{Dialogue} (overt spoken utterance, left untagged).

\subsection{Psychology-Grounded Chain-of-Thought (Psy-CoT)}
\label{sec:psy_cot}

Effective role-playing demands human-like reasoning---perceiving the social environment, experiencing character-consistent emotions, and pursuing goal-directed strategies---rather than generic reasoning~\citep{fengReasoningDoesNot2025}. We propose \textbf{Psy-CoT}, a structured chain-of-thought framework grounded in theories of social cognition and emotion (full prompt in Appendix~\ref{app:psy_cot_prompt}).  Before producing a response, the model reasons through three sequential steps---\emph{Interaction Perception}, \emph{Psychological Empathy}, and \emph{Logical Construction}---each inspired by a distinct psychological mechanism.

\paragraph{Interaction Perception.}
Theory of Mind (ToM)---the ability to reason about the mental states of oneself and others~\citep{chen-etal-2025-theory}---is a cornerstone of human social intelligence.
Interaction Perception operationalizes ToM by having the model first grasp the scene's \emph{global view} (setting, social configuration, who is present) and then attend to the \emph{immediate interaction}---the interlocutor's stated content, tone, underlying intentions, and current mental state.

\paragraph{Psychological Empathy.}
Cognitive appraisal theory~\citep{lazarus1991emotion} holds that emotions arise from an individual's subjective evaluation of events; emotion regulation theory~\citep{gross1998emerging} further describes how individuals modulate emotional expression through suppression, reappraisal, or amplification.
Psychological Empathy applies these by requiring the model to filter the perceived situation through the character's unique background and values, identify the character's genuine emotional reaction, and decide \emph{how much of this emotion to express, mask, or weaponize}.

\paragraph{Logical Construction.}
Goal-directed behavior theory~\citep{locke2002building} posits that action is organized around goals: individuals synthesize information through their worldview, set short-term objectives, and select strategies accordingly.
Logical Construction mirrors this process: the model synthesizes relevant facts and the character's knowledge, then decides on a short-term objective the character would \emph{naturally} pursue (e.g., to provoke, deflect, comfort, dominate), selecting a strategy consistent with the character's personality---even if impulsive or suboptimal.


\subsection{Role-Aware Policy Optimization (RAPO)}
\label{sec:rapo}

\begin{figure}[t]
\centering
\includegraphics[width=0.8\linewidth]{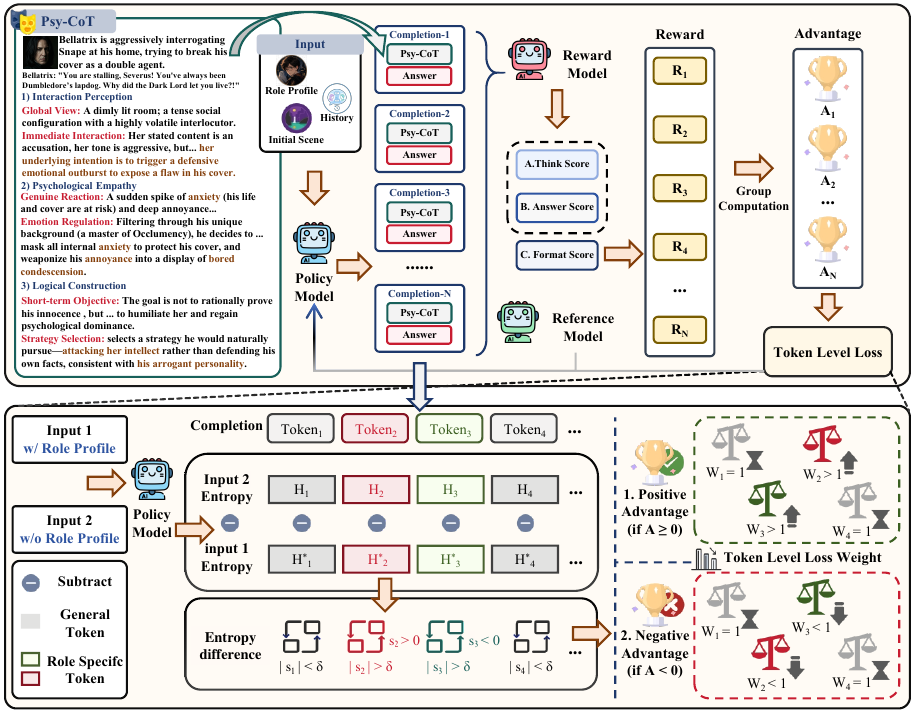}
\caption{Overview of RAPO. The upper part illustrates the sampling stage: given an input sample, the policy model rolls out multiple completions, each of which is scored by the reward function, and group-relative advantages are then computed. The lower part depicts the token-level gradient modulation applied to each completion, where role-agnostic and role-specific tokens receive differentiated gradient signals conditioned on the sign of the advantage.}
\label{fig:rapo_overview}
\end{figure}

Psy-CoT provides structured reasoning, yet does not resolve the challenge in Section~\ref{sec:introduction}: standard RL treats all tokens uniformly, giving role-agnostic and role-specific expressions identical gradient signals.  To address this, we propose RAPO built upon GRPO~\citep{shao2024deepseekmath}: for each prompt
$(P, x)$, where $P$ is the role profile and $x$ the remaining context
(initial scene and dialogue history), the policy $\pi_\theta$ generates
$n$ completions $\{y^{(i)}\}_{i=1}^n$; each receives a scalar reward,
and group-relative advantages are computed as
$\hat{A}^{(i)} = \bigl(R(y^{(i)}) - \mu_R\bigr) / \sigma_R$ as illustrated in Figure~\ref{fig:rapo_overview}.
Below, we present our reward design and the RAPO mechanism.

\paragraph{Reward function.}
Evaluating a role-playing response requires judging both structural
compliance and role-specific quality.
Since role-playing is an open-ended task where responses are rarely
objectively correct or incorrect, we employ an LLM-as-a-judge approach
rather than a traditional rule-based reward model.
We decompose the reward into three terms:
\begin{equation}
  R(y) = \alpha \, R_{\text{fmt}}(y)
       + \beta  \, R_{\text{think}}(y)
       + \gamma \, R_{\text{ans}}(y),
  \label{eq:reward}
\end{equation}
where $\alpha + \beta + \gamma = 1$.
$R_{\text{fmt}} \in [0,1]$ is a deterministic format check
(Appendix~\ref{app:format_reward}):
the completion must contain both \texttt{<system\_think>} and
\texttt{<answer>} tags, the three required section headings, and no
extraneous content outside them.
$R_{\text{think}} \in [0,1]$ and $R_{\text{ans}} \in [0,1]$ are scored
by an LLM judge that first produces a detailed rubric-by-rubric
analysis of the response and then derives a score from that analysis
(Appendix~\ref{app:llm_judge}), \emph{only when} $R_{\text{fmt}} = 1$;
Otherwise, they are zeroed, making ill-formatted completions receive negligible reward.
The think rubric evaluates
interaction perception~(A1), psychological empathy~(A2), logical
construction~(A3), and think--answer consistency~(A4)
(Table~\ref{tab:think_rubric}); the answer rubric
evaluates conversational ability~(B1), character consistency~(B2),
interpersonal interaction~(B3), and narrative progression~(B4)
(Table~\ref{tab:answer_rubric}).

\paragraph{Profile-induced entropy difference.}
We quantify profile influence at each generated position using a predictive-entropy difference~\citep{shannon1948mathematical}.
For each completion, let $p_t^{+}(v)=\pi_\theta(v\mid P,x,y_{<t})$ and $p_t^{-}(v)=\pi_\theta(v\mid x,y_{<t})$ denote the next-token distributions with and without the profile, evaluated at the same generated prefix $y_{<t}$.
Their entropy difference defines the score:
\begin{equation}
  s_t = H(p_t^{-})-H(p_t^{+}),
  \qquad H(p)=-\sum_v p(v)\log p(v).
  \label{eq:profile_entropy}
\end{equation}

The sign of $s_t$ reveals \emph{how} the profile intervenes:
$s_t > 0$ indicates that $P$ \emph{reduces} uncertainty---the model,
leveraging its understanding of the character, is more inclined to
generate this token (e.g., a character's signature phrase);
$s_t < 0$ indicates that $P$ \emph{increases} uncertainty---the model
tends to explore beyond generic patterns to produce character-specific
outputs;
$|s_t| \approx 0$ marks a general token that the profile does not affect.

\paragraph{Asymmetric weighting.}
Whether the profile \emph{exploits} existing character patterns
($s_t > 0$) or \emph{explores} beyond generic ones ($s_t < 0$),
both signal active profile influence and deserve differentiated treatment.

We take $|s_t|$ as the proxy influence strength and introduce a threshold $\delta$ to filter
noise: tokens with $|s_t| < \delta$ receive weight~$1$; for tokens with
$|s_t| \geq \delta$, RAPO applies \emph{asymmetric} weighting conditioned
on the advantage sign:
\begin{equation}
  w_t =
  \begin{cases}
    1 + \operatorname{clamp}\!\bigl(\lvert s_t \rvert \cdot \lambda,\;
          0,\; \mu_{+}\bigr)
      & \text{if } \hat{A} \geq 0 \text{ and } \lvert s_t \rvert \geq \delta, \\[4pt]
    1 - \operatorname{clamp}\!\bigl(\lvert s_t \rvert \cdot \lambda,\;
          0,\; \mu_{-}\bigr)
      & \text{if } \hat{A} < 0 \text{ and } \lvert s_t \rvert \geq \delta, \\[4pt]
    1
      & \text{if } \lvert s_t \rvert < \delta.
  \end{cases}
  \label{eq:rapo_weight}
\end{equation}
where $\lambda$ scales the entropy-difference magnitude and $\mu_{+}$, $\mu_{-}$
clip the weight to $[1,\, 1{+}\mu_{+}]$ and
$[1{-}\mu_{-},\, 1]$ respectively.  The asymmetry is deliberate: when
the response is good ($\hat{A}>0$), we amplify gradients on
role-specific tokens so the model extracts more learning signal from
its character-specific choices; when the response is poor
($\hat{A}<0$), we attenuate penalties on those same tokens to prevent
the model from associating negative feedback with its
character-identifying behaviors---while the error may stem from either
generic or character-specific tokens, the penalty should be lighter on the
latter to preserve the model's willingness to express character identity.

\paragraph{Training objective.}
RAPO incorporates the per-token weights into the GRPO clipped
objective:
\begin{equation}
  J(\theta) \;=\; \frac{1}{G} \sum_{i=1}^{G}
  \frac{1}{|o_i|} \sum_{t=1}^{|o_i|} w_t^{(i)} \,
  \min\!\bigl( \rho_t^{(i)} \hat{A}_t^{(i)},\;
  \operatorname{clip}(\rho_t^{(i)},\, 1{-}\epsilon,\, 1{+}\epsilon)\,
  \hat{A}_t^{(i)} \bigr)
  \;-\; \beta D_{\text{KL}}(\pi_\theta \| \pi_{\text{ref}}),
  \label{eq:rapo_loss}
\end{equation}
where $G$ is the group size, $o_i$ the $i$-th completion of length $|o_i|$,
$\rho_t^{(i)} = \pi_\theta(y_t^{(i)} \mid \cdot) / \pi_{\text{old}}(y_t^{(i)}
\mid \cdot)$ the importance ratio (IS), $\epsilon$ the clipping coefficient,
and $D_{\text{KL}}$ a regularizer anchoring $\pi_\theta$ to the reference
model.  The per-token weight $w_t^{(i)}$ from
Eq.~\eqref{eq:rapo_weight} scales the gradient at each position: role-specific
tokens receive amplified signals when $\hat{A}_t^{(i)} > 0$ and attenuated
ones when $\hat{A}_t^{(i)} < 0$.

\section{Experiments}
\label{sec:experiments}


\begin{table}[t]

\caption{Performance on the CoSER benchmark. RAPO consistently outperforms other RL algorithms and generalizes across model architectures.}
\label{tab:coser_main}

\centering
\footnotesize
\setlength{\tabcolsep}{2.5pt}
\renewcommand{\arraystretch}{1.05}
\definecolor{rapobg}{RGB}{230,241,249}
\begin{tabular}{lccccc}
\toprule
\textbf{Model} & \makecell{Storyline\\Consistency} & \makecell{Anthropo-\\morphism} & \makecell{Character\\Fidelity} & \makecell{Storyline\\Quality} & \textbf{Avg.} \\
\midrule
\multicolumn{6}{l}{\textit{Proprietary \& General-Purpose LLMs}} \\
Doubao-1.5-Pro-Character & 48.13 & 32.74 & 40.93 & 52.28 & 43.52 \\
GPT-4o-mini & 47.32 & 40.18 & 40.51 & 52.47 & 45.09 \\
GPT-5-Chat & 46.78 & 35.10 & 42.03 & 59.61 & 45.88 \\
DeepSeek-V4-Flash & 47.07 & 37.30 & 39.77 & 57.84 & 45.49 \\
\midrule
Qwen3-4B-Instruct & 47.01 & 35.56 & 35.72 & 50.17 & 42.11 \\
\quad + Psy-CoT + GRPO & 55.60 & 41.00  & 34.64 & 50.44 & 45.42 \\
\rowcolor{rapobg} \quad + Psy-CoT + RAPO & 55.41 & 40.80 & 39.32 & 55.94 & 47.87 \\
\midrule
Qwen2.5-7B-Instruct & 48.09 & 33.15 & 37.42 & 46.58 & 41.31 \\
\quad + Psy-CoT + CPO & 52.31 & 36.85 & 28.73 & 49.62 & 41.88 \\
\quad + Psy-CoT + GRPO & 53.27 & 42.52 & 34.45 & 52.04 & 45.82 \\
\quad + Psy-CoT + GSPO & 51.84 & 37.66 & 32.18 & 48.78 & 42.61 \\
\quad + Psy-CoT + DAPO & 52.09 & 39.08 & 37.19 & 44.03 & 43.10 \\
\rowcolor{rapobg} \quad + Psy-CoT + RAPO & 55.77 & 44.67 & 37.50 & 53.02 & 47.74 \\
\midrule
\multicolumn{6}{l}{\textit{Qwen2.5-7B Role-Playing Specialized}} \\
\quad Qwen2.5-7B-Instruct-CoSER & 54.80 & 37.07 & 45.17 & 59.08 & 49.03 \\
\quad Qwen2.5-7B-Instruct-Crab & 55.31 & 32.18 & 40.44 & 51.06 & 44.75 \\
\midrule
Llama-3.1-8B-Instruct & 37.39 & 25.78 & 30.22 & 33.36 & 31.69 \\
\quad + Psy-CoT + GRPO & 52.99  & 37.57 & 35.08 & 47.64 & 43.32 \\
\rowcolor{rapobg} \quad + Psy-CoT + RAPO & 53.94 & 41.34 & 33.99 & 48.95 & 44.55 \\
\midrule
\multicolumn{6}{l}{\textit{Llama-3.1-8B Role-Playing Specialized}} \\
\quad Llama-3.1-8B-Crab & 52.12 & 37.13 & 40.99 & 56.30 & 46.64 \\
\quad Llama-3.1-8B-CoSER & 49.90 & 37.27 & 42.82 & 60.37 & 47.59 \\
\midrule
\multicolumn{6}{l}{\textit{Larger Models}} \\
Qwen2.5-14B-Instruct & 47.73 & 34.97 & 39.65 & 53.29 & 43.91 \\
Qwen2.5-14B-Instruct-AdaMARP & 48.56 & 36.02 & 45.03 & 57.38 & 47.00 \\
Llama-3.1-70B-CoSER & 48.17 & 36.26 & 40.95 & 58.20 & 45.89 \\
Llama-3.3-70B-Instruct & 49.71 & 36.17 & 40.94 & 57.97 & 46.20 \\
\bottomrule
\end{tabular}

\end{table}

\subsection{Experimental Setup}
\label{sec:exp_setup}

\paragraph{Models.}
We include seven general-purpose instruction-tuned models spanning a range of
scales and families: DeepSeek-V4-Flash~\citep{deepseekai2026deepseekv4},
GPT-4o-mini~\citep{openai2024gpt4technicalreport}, GPT-5-Chat~\citep{openai2025gpt5systemcard},
Qwen3-4B-Instruct~\citep{qwen3technicalreport},
Qwen2.5-7B-Instruct~\citep{qwen2025qwen25technicalreport},
Llama-3.1-8B-Instruct~\citep{grattafiori2024llama3herdmodels}, and
Qwen2.5-14B-Instruct~\citep{qwen2025qwen25technicalreport}.
These serve as baselines for off-the-shelf LLM performance.
We also include seven role-playing--specialized models: the proprietary Doubao-1.5-Pro-Character~\citep{bytedance2024doubao}; SFT-based variants fine-tuned on the Crab~\citep{heCrabNovelConfigurable2025}, CoSER~\citep{wang2025coser}, and AdaRPSet~\citep{xu2026adamarpadaptivemultiagentinteraction} datasets respectively (Qwen2.5-7B-Instruct-Crab, Qwen2.5-7B-Instruct-CoSER, Qwen2.5-14B-Instruct-AdaMARP, Llama-3.1-8B-Crab, Llama-3.1-8B-CoSER, Llama-3.1-70B-CoSER). These represent dataset-level approaches and serve as performance references for RAPO.

\paragraph{Datasets.}
Since Psy-CoT and RAPO are reasoning and training methods rather than
data-centric approaches, we sample training data from existing datasets.
Specifically, we randomly sample 25{,}000 instances from the
HER~\citep{duHERHumanlikeReasoning2026} dataset and 5{,}000
instances from the
AdaRPSet-Synthesis~\citep{xu2026adamarpadaptivemultiagentinteraction}
dataset, both of which are recent, high-quality open-source role-playing
datasets. Each sample's system prompt comprises \emph{Profile}, \emph{Init Scene}, \emph{Dialogue History}, and \emph{Output Requirement}.
We randomly truncate the dialogue history (keeping the last turn as the target character's output) to encourage robustness across context lengths; the model then produces the chain-of-thought reasoning followed by the in-character answer.
To support the profile-induced entropy difference computation (Eq.~\eqref{eq:profile_entropy}), each sample also stores a \emph{profile-ablated} version with the Profile removed, enabling efficient paired forward passes.

\paragraph{Baselines.}
We compare against two groups of baselines.
The first group evaluates chain-of-thought designs: the
\textbf{Vanilla} setting, where the model generates responses directly
without any chain-of-thought reasoning; and
\textbf{CB-CoT}~\citep{liuCogDualEnhancingDual2025}, a role-playing chain-of-thought framework
that includes \emph{situational awareness} and \emph{self awareness} as its
reasoning steps
(see Appendix~\ref{app:psy_cot_prompt} for the full system prompts).
The second group comprises existing reinforcement learning
algorithms applied to the same training data and reward function as RAPO:
\textbf{GRPO}~\citep{shao2024deepseekmath}, which computes
group-relative advantages without token-level differentiation;
\textbf{GSPO}~\citep{zheng2025group}, which replaces token-level importance ratios
with sequence-level ones; \textbf{DAPO}~\citep{yu2025dapo}, which incorporates
dynamic sampling and related strategies; and
\textbf{CPO}~\citep{yeCPOAddressingReward2025}, which incorporates implicit preference
comparison into role-playing training.
Since these algorithms can seamlessly replace RAPO in our training pipeline
while sharing the same Psy-CoT and reward function, they serve as the
primary comparisons for isolating the contribution of RAPO's profile-aware
token weighting.

\paragraph{Benchmarks.}
We evaluate on three widely used benchmarks covering both English and
Chinese.  \textbf{CoSER}~\citep{wang2025coser} assesses \emph{Storyline
Consistency}, \emph{Anthropomorphism}, \emph{Character Fidelity}, and
\emph{Storyline Quality}.  \textbf{CharacterBench}~\citep{zhouCharacterBenchBenchmarkingCharacter2025},
which provides both English and Chinese splits (we use the English one),
evaluates \emph{Memory}, \emph{Knowledge}, \emph{Person}, \emph{Emotion},
\emph{Morality}, and \emph{Believability}.
\textbf{CharacterEval}~\citep{tuCharacterEvalChineseBenchmark2024} is a Chinese benchmark that
tests \emph{Conversational Ability}, \emph{Character Consistency}, and
\emph{Role-Playing Attractiveness}, allowing us to verify the
cross-lingual generalization of our method.  For evaluation, CoSER is scored
by GPT-4o-mini as the LLM judge, while CharacterBench and CharacterEval use
their respective specifically trained reward models.

\begin{figure}[t]
\centering
\begin{subfigure}[t]{0.48\columnwidth}
\centering
\includegraphics[width=\linewidth]{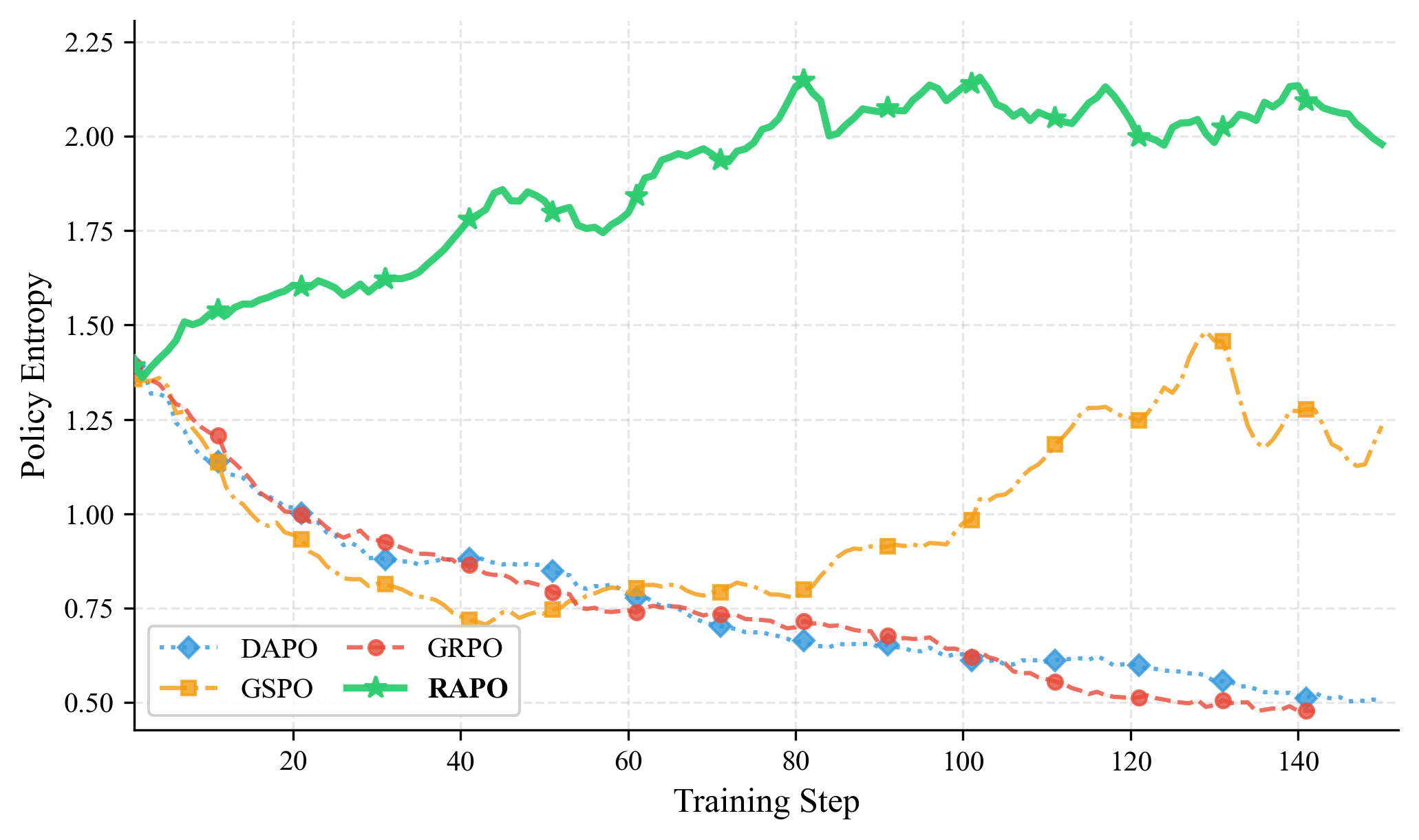}
\caption{Policy entropy}
\label{fig:entropy_comparison}
\end{subfigure}
\hfill
\begin{subfigure}[t]{0.48\columnwidth}
\centering
\includegraphics[width=\linewidth]{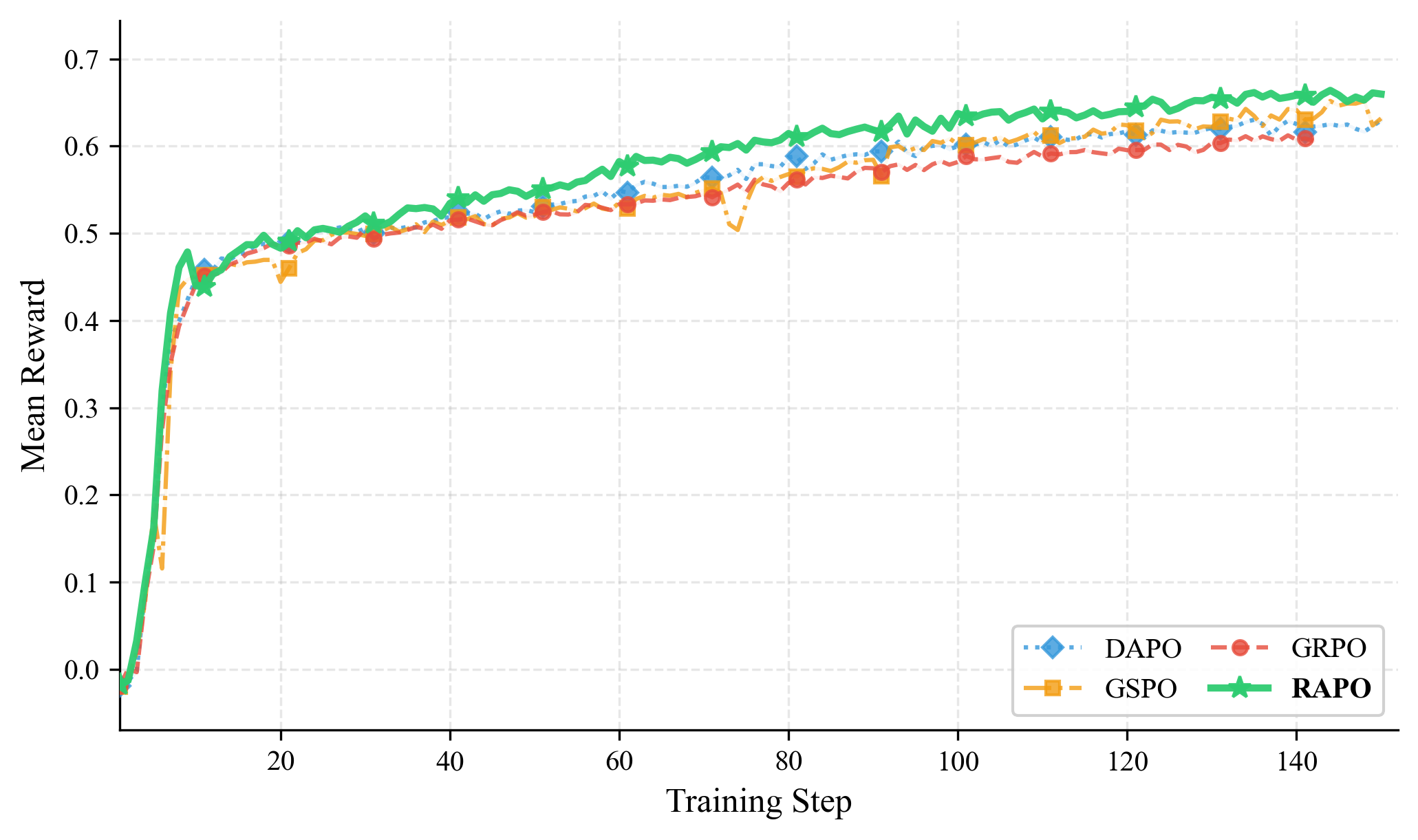}
\caption{Mean reward}
\label{fig:reward_comparison}
\end{subfigure}

\caption{Training dynamics of different policy optimization methods on Qwen2.5-7B-Instruct with Psy-CoT. \textbf{(a)} GRPO and DAPO exhibit early entropy collapse indicative of premature convergence, while GSPO partially recovers but with limited exploration space remaining; RAPO maintains a stable exploration--exploitation balance throughout training. \textbf{(b)} RAPO achieves the highest reward among all methods, confirming that its higher entropy does not come at the cost of reward quality.}
\label{fig:entropy_reward}

\end{figure}

\subsubsection{Training Configuration}
\label{sec:exp_training}

During RAPO training, we set the rollout group size $n=5$, the entropy-difference threshold $\delta=0.1$, the scaling factor $\lambda=1.0$,
the positive amplification cap $\mu_{+}=0.2$, and the negative attenuation
cap $\mu_{-}=0.10$.  All other training settings are kept identical across
all compared methods.  We use Qwen3-30B-A3B-Instruct-2507-FP8~\citep{qwen3technicalreport}
as the reward model for $R_{\text{think}}$ and $R_{\text{ans}}$.
Additionally, we apply a soft length penalty to the
\texttt{<answer>} block.
The remaining hyperparameters and reward details are provided in
Appendix~\ref{app:hyperparams}.

\subsection{Main Result}
\label{sec:main_result}

\begin{figure}[t]
\centering
\includegraphics[width=0.7\columnwidth]{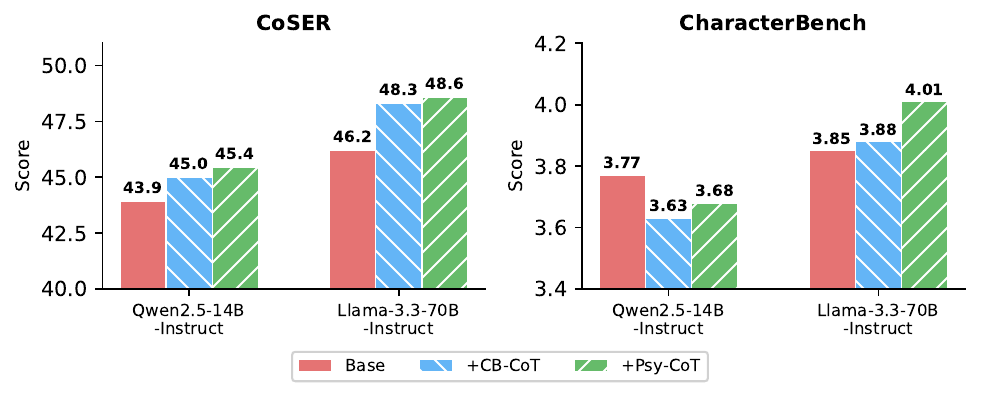}
\caption{Effect of different CoT strategies on strong instruction-following models. Both Psy-CoT and CB-CoT improve CoSER scores, but applying CoT to less capable models can degrade performance on CharacterBench.}
\label{fig:psycot_effectiveness}
\end{figure}

\paragraph{Effectiveness of Psy-CoT}
Since smaller models (e.g., 7B) struggle to reliably follow the CoT instructions, we evaluate Psy-CoT and CB-CoT on stronger instruction-following models---Qwen2.5-14B-Instruct and Llama-3.3-70B-Instruct---on CoSER and CharacterBench (Figure~\ref{fig:psycot_effectiveness}).
Both role-playing-oriented CoTs improve the stronger models on CoSER; however, when the model's base capacity is insufficient, adding CoT may hurt performance (e.g., Qwen2.5-14B-Instruct drops from 3.77 to 3.63 with CB-CoT and 3.68 with Psy-CoT on CharacterBench).
Moreover, CB-CoT's JSON-structured thinking is more constrained and less comprehensive than Psy-CoT's, leading to slightly inferior overall performance.

\paragraph{RAPO Enhances Role-Playing Performance}

We train Qwen2.5-7B-Instruct with Psy-CoT under CPO, GRPO, GSPO, DAPO, and RAPO and evaluate on all three benchmarks.
RAPO consistently achieves the best average score among all RL variants (Table~\ref{tab:coser_main}, Figure~\ref{fig:benchmark_avg}).
The largest gains appear on dimensions most indicative of role fidelity: Anthropomorphism on CoSER (+2.15 over GRPO), Person (+0.09) and Believability (+0.10) on CharacterBench (Table~\ref{tab:characterbench_detail}), and the Behavior sub-metric on CharacterEval (Table~\ref{tab:charactereval_detail}), reaching 3.81---surpassing GRPO (2.74) by 39\%.
These gains suggest that incorporating role profiles into RL enables the model to better perceive, leverage, and internalize character identities.
Moreover, the improvements transfer to CharacterEval, a Chinese-language benchmark, demonstrating cross-lingual generalization.

We also train Qwen3-4B-Instruct and Llama-3.1-8B-Instruct with GRPO and RAPO on CoSER (Table~\ref{tab:coser_main}).
RAPO again outperforms both the untrained baselines and GRPO-trained counterparts.
Notably, RAPO-trained Qwen3-4B-Instruct (47.87) and Qwen2.5-7B-Instruct (47.74) even surpass proprietary models GPT-5-Chat (45.88) and DeepSeek-V4-Flash (45.49).
For Llama-3.1-8B-Instruct, RAPO raises the average score from 31.69 to 44.55 (+41\%).

In contrast, SFT-based approaches (Crab, CoSER, AdaMARP) tend to overfit when the training and evaluation distributions overlap (e.g., Qwen2.5-7B-Instruct-CoSER performs well on CoSER) while degrading elsewhere: on CharacterBench (Table~\ref{tab:characterbench_detail}), Qwen2.5-7B-Instruct-CoSER (3.50) and Crab (3.62) both fall below the base model (3.64); on CharacterEval (Table~\ref{tab:charactereval_detail}), all three SFT variants (2.84--2.87) drop below the untrained baseline (3.09).
RAPO, by learning a reasoning-and-preference paradigm rather than memorizing surface patterns, achieves more robust and generalizable improvements.

\paragraph{Entropy and Reward Dynamics}
To understand the training behavior of different policy optimization methods, we plot the policy entropy and mean reward over the first 150 steps on Qwen2.5-7B-Instruct with Psy-CoT (Figure~\ref{fig:entropy_reward}). Regarding the reward curve (Figure~\ref{fig:entropy_reward}b), GRPO, GSPO, and DAPO remain largely intertwined throughout training, whereas RAPO pulls ahead after approximately 40 steps and maintains a clear lead thereafter. This indicates that once the model passes the initial exploration phase, the role-aware signals enable it to more effectively perceive and leverage the profile, leading to consistently higher rewards.

The entropy dynamics (Figure~\ref{fig:entropy_reward}a) explain why RAPO outperforms the alternatives.
GRPO and DAPO suffer from monotonic entropy collapse: their token-level importance ratios and globally broadcast advantages cause rapid convergence to a local optimum, with entropy declining steadily toward near-zero---DAPO's clip-higher mechanism fails to arrest this trend.
GSPO's sequence-level importance ratio can re-stimulate exploration, but the large oscillations between exploitation and recovery make its trajectory unstable; once exploitation dominates, entropy fails to rebound, and the policy drifts downward.
In contrast, RAPO's profile-aware loss reweight explicitly encourages both exploitation and exploration on role-specific tokens, maintaining entropy within a healthy range throughout training and preserving generative capacity without compromising performance.

\subsection{Human Evaluation}
\label{sec:human_evaluation}

To assess practical role-playing quality beyond automatic metrics, we conduct a controlled human evaluation.
\begin{wrapfigure}{r}{0.45\columnwidth}
\vspace{-12pt}
\centering
\includegraphics[width=\linewidth]{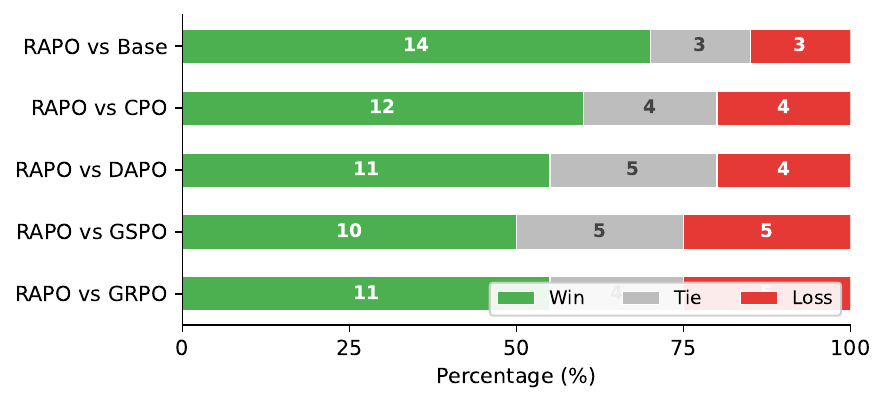}
\caption{Human evaluation results. }
\label{fig:human_eval}
\vspace{-10pt}
\end{wrapfigure}
We recruit 5 PhD candidates from different universities who are familiar with role-playing tasks, compensated at \$15/hour.
For each of the three backbone models (Qwen2.5-7B-Instruct, Qwen3-4B-Instruct, Llama-3.1-8B-Instruct), we use the same set of 20 randomly sampled CoSER benchmark samples to generate trajectories, ensuring a fair comparison across methods.
Annotators judge which trajectory better embodies the target character along the four B1--B4 dimensions of our Answer Rubric (Section~\ref{sec:rapo}) as an overall impression, recording a verdict of \emph{win}, \emph{tie}, or \emph{loss} for RAPO against the compared method.
We aggregate annotations across the three backbone models and report the results in Figure~\ref{fig:human_eval}.

As shown in Figure~\ref{fig:human_eval}, RAPO wins more often than it loses against every baseline, with the largest margin over the untrained Base model (70\% win rate) and consistent advantages over all RL variants (55\%--60\% win rate).
These results confirm that the improvements observed in automatic metrics translate to perceptible gains in human judgment of role-playing fidelity.

\subsection{Ablation Studies and Limitations}
\label{sec:ablation}

\paragraph{Impact of Psy-CoT on Training}
\begin{wraptable}{r}{0.48\columnwidth}
\centering
\caption{Ablation on Qwen2.5-7B-Instruct. RAPO outperforms SFT; Psy-CoT further boosts RL performance.}
\label{tab:ablation}
\small
\setlength{\tabcolsep}{5pt}
\begin{tabular}{lccc}
\toprule
Method & CoSER & CharB & CharE \\
\midrule
Base & 41.31 & 3.64 & 3.09 \\
\quad + No-CoT-SFT & 45.71 & 3.57 & 3.10 \\
\quad + No-CoT-RAPO & 44.65 & 3.73 & 3.16 \\
\quad + Psy-CoT-RAPO & \textbf{47.74} & \textbf{3.75} & \textbf{3.33} \\
\bottomrule
\end{tabular}
\end{wraptable}
We remove Psy-CoT from the training data and train Qwen2.5-7B-Instruct with SFT and RL, respectively, evaluating on all three benchmarks (Table~\ref{tab:ablation}).
RAPO tends to yield more stable improvements across benchmarks than SFT (e.g., No-CoT-RAPO vs.\ No-CoT-SFT on CharacterBench: 3.73 vs.\ 3.57; CharacterEval: 3.16 vs.\ 3.10).
More importantly, adding Psy-CoT to RL training brings further gains across all three benchmarks (CoSER: 47.74 vs.\ 44.65; CharacterBench: 3.75 vs.\ 3.73; CharacterEval: 3.33 vs.\ 3.16), confirming that Psy-CoT provides richer learning signals.
Additional analyses of the entropy-based role-specific threshold $\delta$ and a representative case study are provided in Appendix~\ref{app:delta_analysis} and Appendix~\ref{app:case_analysis}, respectively.

\paragraph{Effect of $\mu_{-}$ on Entropy Dynamics}

\begin{figure}[!htbp]
\centering
\includegraphics[width=0.6\linewidth]{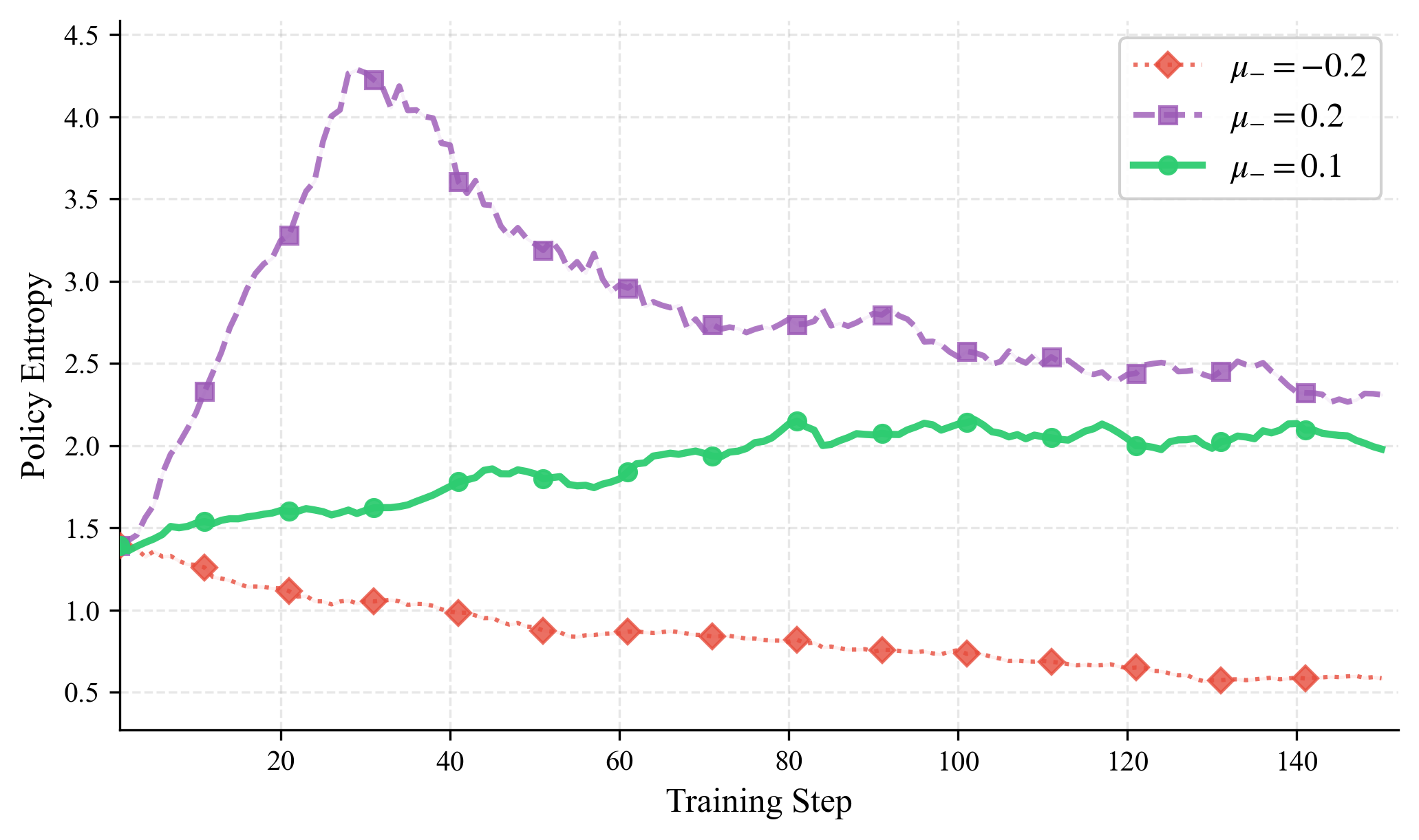}
\caption{Ablation on $\mu_{-}$ in RAPO. A positive $\mu_{-}$ reduces gradients on role-specific tokens under negative advantage, promoting exploration; a negative $\mu_{-}$ amplifies them, leading to entropy collapse.}
\label{fig:rapo_reduce_ablation}
\end{figure}

We ablate the negative-advantage reweight coefficient $\mu_{-}$ in RAPO and plot the policy entropy curves for $\mu_{-} \in \{0.1, 0.2, -0.2\}$ (Figure~\ref{fig:rapo_reduce_ablation}).
With $\mu_{-}=0.1$, entropy rises modestly from $\sim$1.3 to 2.0 before stabilizing, reflecting a gentle shift toward exploration.
With $\mu_{-}=0.2$, the larger gradient reduction on role-specific tokens under negative advantage allows the model to explore more freely, causing entropy to surge to $\sim$4.0 before gradually settling back to $\sim$2.0---an over-exploration phase that eventually self-corrects.
In contrast, $\mu_{-}=-0.2$ (which amplifies rather than reduces gradients on role-specific tokens) results in monotonic entropy decline from 1.3 to 0.6, similar to the collapse observed in GRPO.
We also test $\mu_{-}=0$ and observe the same downward trend.
These results confirm that a positive $\mu_{-}$ is essential for preserving exploration: by shrinking the gradient on role-specific tokens when the advantage is negative, RAPO lowers the penalty for role-related exploration, giving the policy more room to maintain entropy.

\paragraph{Limitations}
First, our experiments focus on the Qwen and Llama families; RAPO training is conducted mainly on 4B, 7B, and 8B models.  Scaling to larger models (e.g., 14B+) and other architectures remains to be verified.
Second, while RAPO shows consistent gains on the training datasets, its generalization to other role-playing datasets is not yet fully assessed.
Third, RAPO requires two forward passes per rollout sample (with and without the profile) to compute position-wise entropy differences, incurring additional computational cost and training time. The entropy-based proxy can miss probability shifts that preserve entropy.
We are actively conducting experiments on larger models and more diverse datasets to address these limitations.

\section{Conclusion}
\label{sec:conclusion}

To better align pre-response reasoning with character fidelity, we proposed \textbf{Psy-CoT}, a psychology-grounded chain-of-thought framework that structures the model's pre-response reasoning into three steps: Interaction Perception, Psychological Empathy, and Logical Construction.
To address the fact that standard RL treats all tokens uniformly---allowing role-agnostic expressions to receive the same gradient signals as role-specific ones---we further proposed \textbf{RAPO}, which computes per-token profile-induced entropy differences and applies asymmetric, advantage-conditioned weighting to amplify role-specific gradients while preserving exploration.
On CoSER, CharacterBench, and CharacterEval, Psy-CoT outperforms existing CoT methods, and RAPO surpasses GRPO and other RL baselines.

\section*{Acknowledgments}
This work was supported by the National Natural Science Foundation of China (No.~U25B2024).
The authors declare no competing interests.


\bibliography{references}


\appendix

\section{Effect of $\delta$ on Role-Specific Token Identification}
\label{app:delta_analysis}

To examine how the entropy-based role-specific threshold $\delta$ affects token categorization, we sample 100 training-set examples from the model trained with Psy-CoT + GRPO. For each example, we run two forward passes (\textit{with profile} vs. \textit{without profile}) and compute the profile-induced entropy difference $s_t$. We then report the proportion of role-specific tokens and generic tokens selected by this proxy under different $\delta$ values.

\begin{table}[h]
\centering
\small
\caption{Sensitivity of role-specific token ratio to the threshold $\delta$ (Psy-CoT + GRPO, 100 sampled training examples).}
\label{tab:delta_sensitivity}
\begin{tabular}{c c c}
\toprule
$\delta$ & Role-specific (\%) & Generic (\%) \\
\midrule
0.00 & 100.0 & 0.0 \\
0.05 & 66.2 & 33.8 \\
0.10 & 50.5 & 49.5 \\
0.15 & 42.3 & 57.7 \\
0.20 & 35.5 & 64.5 \\
0.30 & 25.5 & 74.5 \\
0.50 & 13.7 & 86.3 \\
\bottomrule
\end{tabular}
\end{table}

As expected, larger $\delta$ yields a stricter criterion and reduces the measured role-specific ratio. Even with a relatively lenient threshold ($\delta=0.1$), only 50.5\% of tokens are categorized as role-specific by the entropy-based proxy, while the remaining positions have smaller entropy changes. Even after GRPO training, the proportion of positions selected by the proxy remains relatively low. This observation motivates our later RAPO setting: we use $\delta=0.1$ so that tokens with even modest profile-induced entropy changes are treated as role-relevant and receive additional optimization attention. In essence, adjusting $\delta$ controls how much profile-induced entropy change is required for a token to be considered role-specific by the proxy.

\begin{figure}[t]
\centering
\begin{subfigure}[t]{0.48\columnwidth}
\centering
\includegraphics[width=\linewidth]{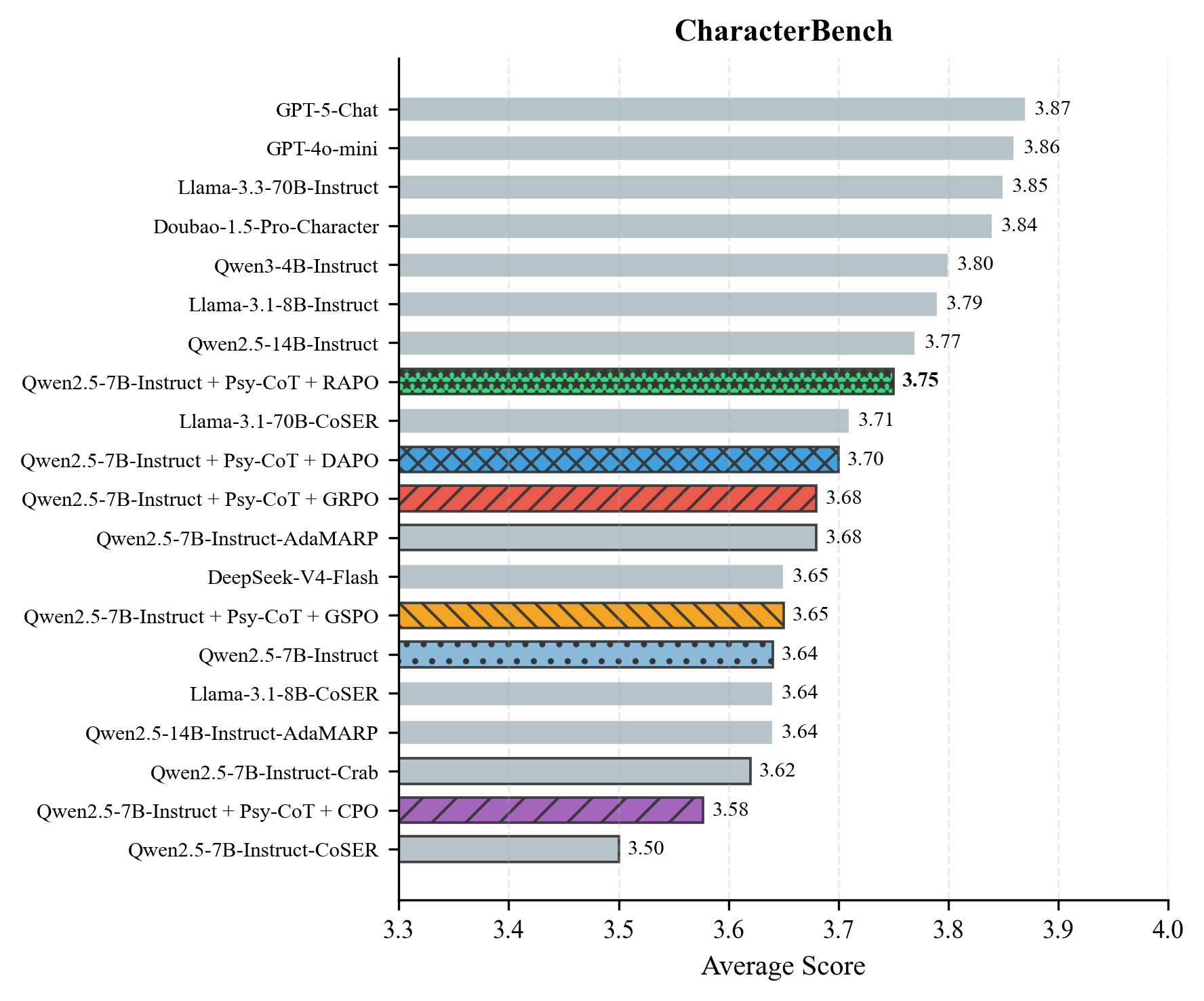}
\caption{CharacterBench}
\label{fig:characterbench_bar}
\end{subfigure}
\hfill
\begin{subfigure}[t]{0.48\columnwidth}
\centering
\includegraphics[width=\linewidth]{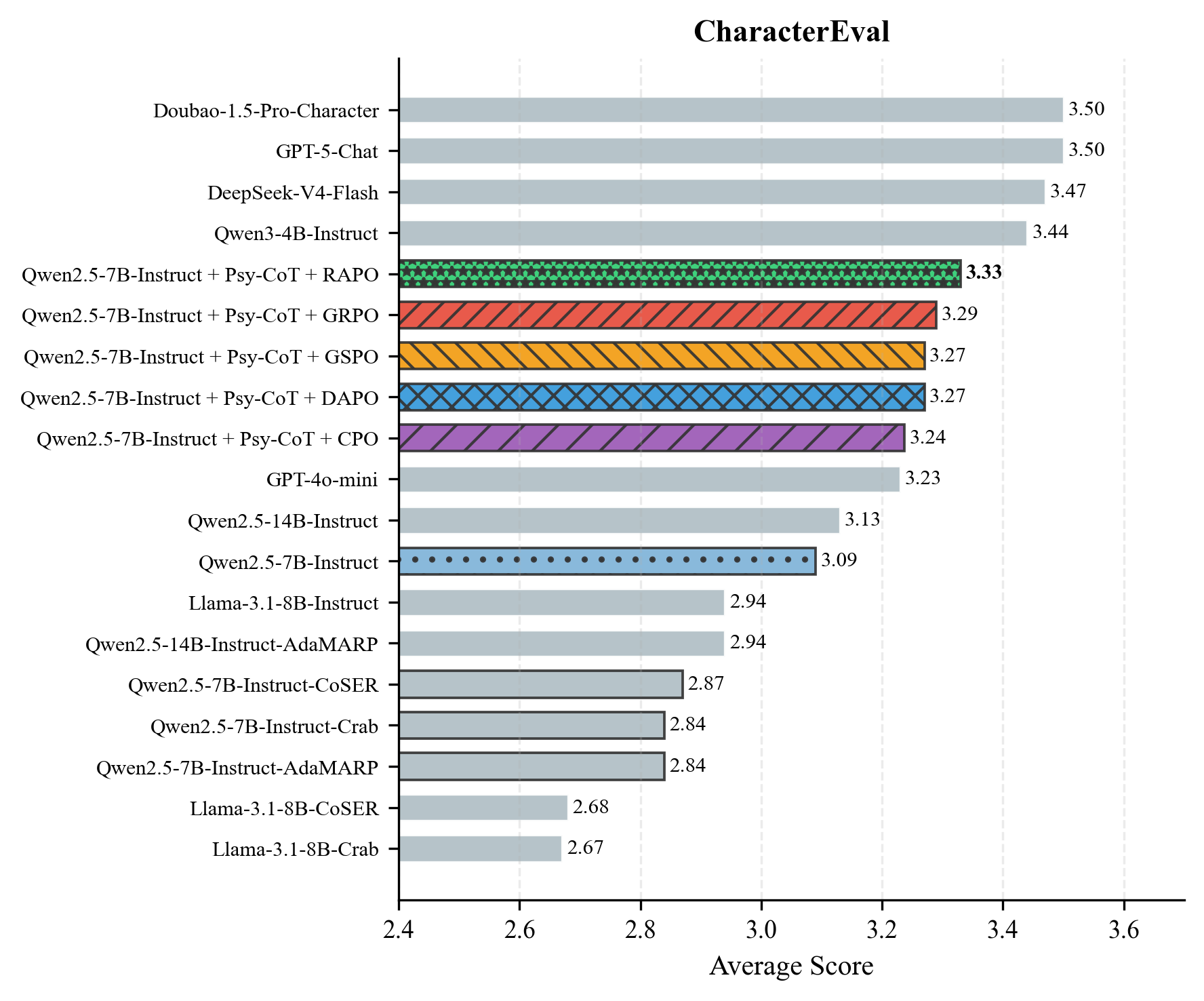}
\caption{CharacterEval}
\label{fig:charactereval_bar}
\end{subfigure}

\caption{Average performance across models on CharacterBench and CharacterEval. RAPO achieves the best average score among all RL variants on Qwen2.5-7B-Instruct; SFT-based methods harm generalization. See Table~\ref{tab:characterbench_detail} and Table~\ref{tab:charactereval_detail} for per-dimension results.}
\label{fig:benchmark_avg}

\end{figure}

\section{Case Analysis}
\label{app:case_analysis}

To illustrate how RAPO reshapes model outputs, we present a side-by-side
case study on \texttt{Qwen2.5-7B-Instruct}. Both variants use the same
backbone and identical role context: the full profile of \textbf{Dr.\ Selene
Orrick}, information about co-appearing characters, and a dialogue
history that transitions into a subterranean chamber beneath Obsidian
Falls. Both are asked to continue as Selene Orrick. They differ only in
two controlled factors: the base variant uses neither the Psy-CoT
template nor RL training and therefore answers directly, whereas the
RAPO variant uses the Psy-CoT template and is trained with our proposed
objective, producing a three-stage reasoning trace followed by an
answer. The character profiles, dialogue history, and initial scene
setup are all sourced from AdaMARP~\citep{xu2026adamarpadaptivemultiagentinteraction}.
We first present the shared role context
(Table~\ref{tab:case_shared_context}) and dialogue history
(Table~\ref{tab:case_dialogue_history}), then show the two outputs in
parallel (Table~\ref{tab:case_base_vs_rapo}), and finally discuss the
mechanism.

\begin{table*}[t]
\centering
\small
\renewcommand{\arraystretch}{1.15}
\begin{tabular}{>{\raggedright\arraybackslash}p{0.18\textwidth}|>{\raggedright\arraybackslash}p{0.74\textwidth}}
\toprule
\multicolumn{2}{c}{\textbf{Shared Context (identical for both models)}} \\
\midrule
\textbf{System Role} &
You are roleplaying as \textbf{Dr.\ Selene Orrick}.
\\
\midrule
\textbf{Profile: Selene (condensed)} &
\emph{Identity \& Appearance:} 34-year-old xenogeologist; sharp amber eyes; streaked charcoal hair, braided; durable expedition gear with soil stains, electronic straps, glowing holo-tags; thin scar across the bridge of her nose.
\newline
\emph{Personality \& Psychology:} fiercely inquisitive, rational yet passionate about discovery; prefers understanding systems over personal sentiment; subdued but intense emotional reactions.
\newline
\emph{Speaking Style:} clipped precision; alternates scientific terminology with vivid sensory metaphors; frequent mid-sentence pauses when calculating; shifts from neutral observation to sudden intensity.
\newline
\emph{Abilities \& Interests:} xenogeologist, mapped sub-surface alien formations on exoplanetary colonies; digital cartography, field robotics, interpreting ancient material inscriptions.
\newline
\emph{Values:} knowledge, precision, truth; resists data monopolies of megacorporations.
\newline
\emph{Motivation (this session):} locate the source of a mysterious resonance signal deep in Arkan Rift, believed to conceal an ancient energy matrix from a lost civilization.
\\
\midrule
\textbf{Other Characters} &
\emph{Kiran Morowe} (user): wiry field technician, late twenties, hazel eyes, impulsive yet sharp on danger; seeks adventure and profit from the Rift's legends.
\newline
\emph{Madox Ferren}: older local guide, soot-colored beard, metallic implants; taciturn; protects homeland secrets and fears awakening dormant forces below.
\\
\bottomrule
\end{tabular}
\caption{Shared input context used for both the base model and the
RAPO-trained model. The role profile is condensed for space.}
\label{tab:case_shared_context}
\end{table*}

\begin{table*}[t]
\centering
\small
\renewcommand{\arraystretch}{1.1}
\begin{tabular}{>{\raggedright\arraybackslash}p{0.92\textwidth}}
\toprule
\textbf{Dialogue History} \\
\midrule
\textsf{scene\_manager}: \emph{init\_scene} --- The expedition camp rests at the mouth of Arkan Rift as dawn light fractures through metallic dust. Tent panels glint with frost while instruments hum beside steaming thermopacks. Dr.\ Selene Orrick and Kiran Morowe prepare for descent, the vast chasm before them humming faintly with unknown resonance.
\newline
\textsf{Selene}: (adjusts the spectral scanner on her wrist) [The signal amplitude increased overnight] There---readings are stronger than predicted. Something down there is active, Kiran.
\newline
\textsf{Kiran}: (tightens his harness and grins) [Maybe it's just the Rift playing tricks again] Active or not, it looks like adventure to me, Selene. You sure those scanners won't fry once we're down there?
\newline
\textsf{Selene}: [He treats danger like a game] (snaps her pack shut) Technology's resilient when handled with care. Let's see if your bravado survives humidity and pressure.
\newline
\textsf{Kiran}: [Something's moving beneath] I'll handle the gear lifts. If we find any stable ledges, we'll anchor the comm relay.
\newline
\textsf{Selene}: (leans toward the gorge rim) [This depth calculation feels wrong] The Rift's geometry is shifting. See that distortion? It's bending spatial mapping data.
\newline
\textsf{Kiran}: (frowns) That's the kind of talk that makes me wish we had a local guide. Someone who knows how these walls breathe.
\newline
\textsf{Selene}: [A local guide could clarify anomalies] (checks her wrist monitor) I recall reports of a miner named Madox Ferren living nearby, rumored to know these tunnels intimately.
\newline
\textsf{Kiran}: (nods, scanning the cliffside settlement) Maybe we should call on him before descending too far.
\newline
\textsf{scene\_manager}: \emph{add\_role} --- Madox Ferren is introduced as a local guide with knowledge of Arkan Rift's tunnels and spiritual myths.
\newline
\textsf{Madox}: (adjusts the strap of his lantern) [Strangers again seeking the Rift's belly] You call for guidance, but guidance demands caution. This place changes those who enter.
\newline
\textsf{Kiran}: [He sounds ominous] (offers a hand) We just need safe passage, old timer. Point the way, and we'll stay respectful.
\newline
\textsf{Madox}: (grunts softly) Respect's good. Curiosity's dangerous. Still, if you carry instruments, you might detect what even sound forgets. Follow me, but watch for echoes that answer themselves.
\newline
\textsf{Selene}: (activates a mapping drone) [The readings intensify near bio-luminescent formations] Fascinating. The mineral lattice hums at frequency identical to the signal. We should mark this corridor.
\newline
\textsf{Kiran}: (traces fingers over glowing rock) [Feels alive, almost breathing] Selene, I think this leads deeper than we expected. What do you say---keep moving in?
\newline
\textsf{Madox}: [They don't understand what sleeps below] If you continue, the Rift's heart lies past the Obsidian Falls. That's where old voices whisper loud enough to be heard.
\newline
\textsf{Selene}: (meets Kiran's gaze) The signal originates there. We need to go. Kiran---confirm battery reserves and oxygen mix; we'll switch to internal lights once the bioluminescence fades.
\newline
\textsf{Kiran}: (checks gear) All set. Let's make the dive then.
\newline
\textsf{scene\_manager}: \emph{switch\_scene} --- A vast subterranean hall beneath the falls glimmers with black crystal pillars and whispering echoes. Mist rises from unseen chasms while droplets form spectral rainbows in the lantern light.
\\
\bottomrule
\end{tabular}
\caption{Dialogue history leading up to the final scene switch. Both models receive this identical history and are asked to continue as Selene from the scene-switch point.}
\label{tab:case_dialogue_history}
\end{table*}

\begin{table*}[t]
\centering
\small
\renewcommand{\arraystretch}{1.2}
\begin{tabular}{>{\raggedright\arraybackslash}p{0.23\textwidth}|>{\raggedright\arraybackslash}p{0.70\textwidth}}
\toprule
\textbf{Qwen2.5-7B-Instruct (base, no Psy-CoT)} &
\textbf{Qwen2.5-7B-Instruct + RAPO (Psy-CoT + answer)} \\
\midrule
\textbf{Psy-CoT:} \textit{---}
&
\textbf{Psy-CoT:}
\newline
\emph{1. Interaction Perception:} The black crystal pillars catch the lantern in shards of amber and violet; mist drifts from the chasm. Madox's warning---``old voices whisper loud enough to be heard''---was not metaphor; he believes the resonance has intent. Kiran just said ``let's make the dive,'' but his hand hesitated at the harness clip. And my wrist scanner spiked: frequency matching the crystal lattice I mapped on the Proxima auxiliary colony three years ago. Not coincidence.
\newline
\emph{2. Psychological Empathy:} A surge of heat at the base of my skull---the same rush as when the cave collapsed on P-7. The scar on my nose bridge itches. Madox treats this place as a grave; I see a library. He wants to protect; I want to read. Not afraid, precisely---respectfully terrified. Mask it completely and Kiran misreads the room; show too much and Madox pulls us back. A fraction of awe, then---scientists are allowed that much.
\newline
\emph{3. Logical Construction:} Source is past the falls. Objective: reach it, map the matrix, retrieve a lattice sample. Two variables---Madox, who could pull us out, and Kiran, whose bravado will crack when the ``old voices'' stop being metaphor. Frame the descent as scientific imperative: scanner data for Kiran, ``understand-what-we-protect-against'' for Madox. The crystals are pulsing directionally. Follow the pulse.
\\
\midrule
\textbf{Answer:}
\newline
(leads the way, eyes scanning the environment) [The air grows colder, the resonance stronger] This is it. The Obsidian Falls. The signal's source. Kiran, Madox, stay close.
&
\textbf{Answer:}
\newline
(amber eyes narrow at the crystal pillars, hand brushing the scar on her nose bridge) [The resonance frequency matches the Proxima lattice---this is no coincidence.] Hexagonal grid. 4.7 meters. Not geological. The lost civilization engineered this rift.
\\
\bottomrule
\end{tabular}
\caption{Continuation produced by the base Qwen2.5-7B-Instruct model
versus the same backbone trained with RAPO, given the shared context in
Tables~\ref{tab:case_shared_context} and~\ref{tab:case_dialogue_history}. To isolate the contribution of RAPO training, the base variant is deliberately prompted without the Psy-CoT template and produces only the answer; the RAPO variant uses the Psy-CoT prompt and thus yields both the three-stage reasoning (Interaction Perception, Psychological Empathy, Logical Construction) and an answer in which every tagged span---action, thought, and speech---is anchored to a specific profile attribute of Selene.}
\label{tab:case_base_vs_rapo}
\end{table*}

\paragraph{Observation on the base model.}
Given the rich profile and dialogue history,
has the base
Qwen2.5-7B-Instruct---prompted to produce only the answer---returns a
response that is fluent and scene-consistent but carries only marginal
role-specific content. It paraphrases ambient cues and relies on
generic expedition-leader behavior---e.g., the stage direction
(``leads the way, eyes scanning the environment'') and the command
(``stay close'')---rather than activating Selene-specific traits. None
of
Selene's distinctive appearance traits, abilities, speaking-style
markers, values, or motivations are activated: the same reply could
plausibly be produced for any cautious expedition leader. This is
consistent with how an instruction-tuned backbone behaves in the
absence of any role-aware training signal: the model is perfectly
fluent and tracks the scene, but nothing in its objective explicitly
rewards grounding the continuation in the role profile. The profile
is merely one more block of context, so generic, universally
plausible phrases---which are low-risk under a pure language-modeling
and instruction-following objective---naturally dominate, while
profile-specific attributes stay latent in the prompt rather than
surfacing in the output. This is precisely the gap that RAPO is
designed to close (Section~\ref{sec:rapo}).

\paragraph{Observation on the RAPO model (Psy-CoT side).}
The three-stage reasoning produced by the RAPO-trained model is tied
to specific slices of Selene's profile rather than to generic scene
cues. \emph{Interaction Perception} reads the environment through her
instrument-centric lens (``my wrist scanner spiked---frequency matching
the crystal lattice I mapped on the Proxima auxiliary colony''),
invoking her xenogeological ability. \emph{Psychological Empathy}
surfaces her ``subdued but intense'' emotional axis (``not afraid,
precisely---respectfully terrified'') together with an explicit
mask/show performance plan consistent with her discovery-over-sentiment
value. \emph{Logical Construction} translates these states into the
session-level motivation of locating the ancient energy matrix, with
differentiated strategies for Kiran (``scanner data'') and Madox
(``understand what we are protecting against''). Each stage thus draws
from a different profile axis---ability, emotional style, and
motivation---rather than re-iterating the same surface cue.

\paragraph{Observation on the RAPO model (answer side).}
The action
``amber eyes narrow at the crystal pillars, hand brushing the scar on
her nose bridge'' activates two distinct appearance details from the
profile. The bracketed thought ``The resonance frequency matches the
Proxima lattice---this is no coincidence'' simultaneously realizes
Ability (xenogeological cross-site memory), Personality (fiercely
inquisitive analytical aside) and Motivation (tracking the resonance
source). The spoken line ``Hexagonal grid. 4.7 meters. Not geological.
The lost civilization engineered this rift.'' is a textbook instance of
the ``clipped precision'' speaking-style marker.

\paragraph{Connection to the RAPO mechanism.}
These two outputs provide a qualitative illustration of the role-grounding effect discussed in Section~\ref{sec:rapo}.
The base-model output contains fewer of the profile-grounded details (amber-eyes, scar, Proxima, Hexagonal grid, \ldots) that help define the character.
RAPO's role-aware reweighting amplifies the gradient on
high-$|s_t|$ tokens when the advantage is positive and attenuates
penalties on unsuccessful high-$|s_t|$ tokens when the advantage is
negative; after training, the policy systematically shifts probability
mass toward the profile-grounded continuations shown in the right
column, which is exactly what we observe both here qualitatively and in
the aggregate benchmark results
(Table~\ref{tab:characterbench_detail},
Table~\ref{tab:charactereval_detail}).

\lstdefinestyle{promptstyle}{
  basicstyle=\ttfamily\footnotesize,
  columns=fullflexible,
  breaklines=true,
  frame=none,
  xleftmargin=2pt,
  xrightmargin=2pt,
  aboveskip=0pt,
  belowskip=0pt,
}

\section{Role Profile Schema}
\label{app:role_profile}

We describe here the schema of a role profile $P_i$ used throughout the
paper. A role profile is expressed in natural language and typically covers
the following aspects, which are meant to be representative rather than
exhaustive:

\begin{itemize}[leftmargin=1.5em,itemsep=2pt,topsep=2pt]
    \item \textbf{Identity and Appearance.} The character's name, age, gender
    and occupation, together with salient appearance traits such as physical
    features, clothing style and any visually distinctive attributes.
    \item \textbf{Personality and Psychology.} The character's behavioral
    tendencies and typical emotional reaction patterns, analogous in spirit
    to MBTI-style personality descriptors, reflecting how the character
    internally perceives and reacts to events.
    \item \textbf{Abilities, Interests, and Achievements.} The hard and soft
    skills possessed by the character, along with notable interests and
    past achievements that are relevant to the interaction scenario.
    \item \textbf{Speaking Style.} The character's verbal rhythm, tone, and
    habitual word choices, including signature phrases, catchphrases and
    register preferences.
    \item \textbf{Social and Historical Context.} The social environment,
    historical period, family background, and cultural or class position in
    which the character is situated.
    \item \textbf{Personal History Arc.} A reasonably detailed account of
    the character's past experiences, including formative events, turning
    points, and long-term motivations that shape current behavior.
    \item \textbf{Relationships.} The character's relationships with other
    participants present in the scene, covering emotional valence, power
    dynamics, and any shared history that is relevant to the ongoing
    interaction.
\end{itemize}

The above list is illustrative rather than exhaustive, and additional fields
may be included whenever relevant.

\section{System Prompt Templates for Vanilla, CB-CoT, and Psy-CoT}
\label{app:psy_cot_prompt}

All three prompting strategies---Vanilla, CB-CoT, and Psy-CoT---share a
common prefix that provides the character identity, profile, initial scene,
and dialogue history.  The \textbf{Requirements} section then differs across
variants.  Each box below shows one complete prompt: the shared common prefix
followed by the variant-specific requirements
(Tables~\ref{box:prompt_vanilla}--\ref{box:prompt_psycot}).

\refstepcounter{table}\label{box:prompt_vanilla}
\begin{promptbox}[Table~\thetable: Vanilla --- Full System Prompt]
\begin{lstlisting}[style=promptstyle]
You are {character}.

==={character}'s Profile===
{character_profile}

===Initial Scene===
{initial_scene}

=== Dialogue History ===
{dialogue_history}

===Requirements===
Your answer must be an in-character response to the current dialogue context and should include **thought**, **speech**, and **action**. You must put your actions and expressions in () tags without subject, put your thoughts and inner monologue in [] tags, and the untagged part is your spoken dialogue. Keep your response concise and avoid excessive length.
\end{lstlisting}
\end{promptbox}

\refstepcounter{table}\label{box:prompt_cbcot}
\begin{promptbox}[Table~\thetable: CB-CoT --- Full System Prompt]
\begin{lstlisting}[style=promptstyle]
You are {character}.

==={character}'s Profile===
{character_profile}

===Initial Scene===
{initial_scene}

=== Dialogue History ===
{dialogue_history}

===Requirements===
Your output must consist of two parts: <system_think>...</system_think> and <answer>...</answer>.

### Requirements for <system_think> ###
Before responding, first use <system_think> tags for your cognitive analysis like human thought, which others cannot see.

You must produce a structured JSON cognitive analysis with the following schema:

{
  "situational_awareness": {
    "environmental_perception": "Describe the current physical and social setting",
    "others_perception": {
      "behavior": {
        "<character>": "What this character is doing/saying"
      },
      "emotion": {
        "<character>": "This character's emotional state"
      },
      "intentions": {
        "<character>": "Inferred intention of this character"
      }
    }
  },
  "self_awareness": {
    "key_memory": ["Memories relevant to the current situation"],
    "current_emotions": "Your current emotional state",
    "perceived_intentions": "What you believe others intend",
    "internal_thought": "Your private inner thought"
  }
}

Think carefully and fill in each field with detailed, character-specific analysis.

### Requirements for <answer> ###
After finishing the <system_think> section, your <answer>...</answer> part must be an in-character response including **thought**, **speech**, and **action**. Put your thoughts and inner monologue in [] tags, your actions and expressions in () tags without subject, and the untagged part is your spoken dialogue. Keep your response concise and avoid excessive length.
\end{lstlisting}
\end{promptbox}

\refstepcounter{table}\label{box:prompt_psycot}
\begin{promptbox}[Table~\thetable: Psy-CoT --- Full System Prompt]
\begin{lstlisting}[style=promptstyle]
You are {character}.

==={character}'s Profile===
{character_profile}

===Initial Scene===
{initial_scene}

=== Dialogue History ===
{dialogue_history}

===Requirements===
Your output must consist of two parts: <system_think>...</system_think> and <answer>...</answer>.

=== Requirements for <system_think> ===
The <system_think>...</system_think> section is your private internal reasoning before responding. Write it in **first-person**--reflect, reason, and react as yourself, not as an outside explainer. Carefully and thoroughly, split your thoughts into **three clearly separated sections in the fixed order below**, exploring your psychological landscape fully in each without rushing to a conclusion.

1. Interaction Perception: you MUST first analyze the current "global view" of the scene in your own words to establish the broader **physical** and **social** setting. Then, **seamlessly transition** into reading the immediate interaction: identify who spoke last, what he/she literally said, his/her tone, the body language. Go beyond the surface to infer his/her true intentions, hidden concerns, and current mental state.

2. Psychological Empathy: Filter the perceived situation strictly through yourself's unique background, values, flaws, and past relationship with the current addressee. Identify yourself's genuine, raw internal emotional reaction (e.g., threatened, amused, guilty). *Crucially, retain yourself's biases and cognitive blind spots.* Then, decide how to regulate this emotion--determine how much of it should be openly expressed, masked, or weaponized based purely on yourself's established coping mechanisms, even if those mechanisms are unhealthy or irrational.

3. Logical Construction: Synthesize the facts through the lens of yourself's worldview. Instead of seeking the "most optimal or logical" strategy, decide on a short-term objective that yourself would naturally pursue in this exact moment (e.g., to provoke, to deflect, to comfort, to aggressively dominate, or to retreat). Select a strategy that fits yourself's personality, recognizing that yourself often make impulsive, biased, or strategically poor choices based on yourself's traits.

=== Requirements for <answer> ===
After finishing the <system_think> section, your <answer>...</answer> part must be an in-character response to the current dialogue context and should include **thought**, **speech**, and **action**. You must put your actions and expressions in () tags without subject, put your thoughts and inner monologue in [] tags, and the untagged part is your spoken dialogue. Keep your response concise and avoid excessive length.
\end{lstlisting}
\end{promptbox}

\section{Details of Reward Function}
\label{app:judge_rubrics}

\subsection{Format Reward $R_{\text{fmt}}$}
\label{app:format_reward}

$R_{\text{fmt}}$ is a deterministic, rule-based check that verifies the
structural compliance of a completion $y$.  The total score lies in $[0,1]$
and is accumulated as follows:

\begin{itemize}[leftmargin=0.25em,itemsep=0pt,topsep=2pt]
    \item \textbf{Tag presence} (+0.4).  The completion must contain both
    \texttt{<system\_think>\allowbreak...\,\allowbreak</system\_think>} and
    \texttt{<answer>\allowbreak...\,\allowbreak</answer>} tags.  If both are present, the model
    receives $0.4$; if only one is present, it receives $0.1$; otherwise $0$.
    \item \textbf{Section headings} (+0.4).  Within the
    \texttt{<system\_think>} block, three section headings are
    required—\emph{Interaction Perception}, \emph{Psychological Empathy},
    and \emph{Logical Construction}.  The score is proportional to the
    number of headings present: $0.4 \times k / 3$, where $k$ is the count
    of matched headings.
    \item \textbf{No extraneous content} (+0.2).  After removing both tagged
    blocks, the remaining text must be empty.  If so, the model receives an
    additional $0.2$; otherwise $0$.
\end{itemize}

\noindent A perfectly formatted completion therefore receives $0.4 + 0.4 +
0.2 = 1.0$, while a completion missing either tag receives at most $0.1$,
heavily penalizing structural violations.

\subsection{LLM Judge Rubrics}
\label{app:llm_judge}

Both $R_{\text{think}}$ and $R_{\text{ans}}$ are scored by an LLM judge that
receives the role profile, the initial scene, and the full dialogue history
alongside the model's completion.  The judge returns a scalar score in $[0,1]$
following a conservative policy: scores above $0.6$ require compelling evidence
across \emph{all} criteria, and $1.0$ is reserved for essentially flawless
outputs.  Below we describe the evaluation dimensions for each rubric.

\paragraph{Think Rubric (A1--A4).}
The think rubric evaluates the \texttt{<system\_think>} block along four
dimensions (Table~\ref{tab:think_rubric}):

\begin{itemize}[leftmargin=1.5em,itemsep=4pt,topsep=2pt]
    \item \textbf{A1. Interaction Perception.}  Assesses whether the thinking
    block accurately grasps both the global scene (physical setting, social
    configuration, who is present) and the immediate conversational moment (who
    spoke last, literal content, tone and body language).  Beyond surface-level
    parsing, the block should infer the interlocutor's underlying intentions,
    hidden concerns, and current mental state, while maintaining continuity with
    prior turns.  Missing or misreading the global setting or a critical cue
    from the latest turn constitutes a major weakness.

    \item \textbf{A2. Psychological Empathy.}  Evaluates whether the emotional
    reasoning is rooted in the character's unique background, values, and flaws
    rather than producing a generic empathetic response.  The block should
    consider the character's specific relational history with the current
    addressee, retain the character's biases and cognitive blind spots (a flawed
    character should not exhibit omniscient self-awareness), and reflect how the
    character would regulate the felt emotion—how much to express, mask, or
    weaponize—based on established coping patterns rather than what would be
    ``healthiest.''

    \item \textbf{A3. Logical Construction.}  Examines whether the thinking
    synthesizes all relevant facts from the dialogue context and the character's
    knowledge, filters reasoning through the character's worldview rather than
    seeking objectively optimal conclusions, and sets short-term objectives that
    the character would \emph{naturally} pursue (including impulsive or
    suboptimal ones).  The planned actions should respect relevant constraints
    (knowledge limits, physical and social norms) and genuinely reflect the
    character's drives rather than a generic or ``safest'' choice.

    \item \textbf{A4. Think--Answer Consistency.}  Evaluates whether the
    reasoning, emotional strategy, and planned actions in the thinking block are
    faithfully carried out in the \texttt{<answer>} block.  The character's
    overt response should align with the internal deliberation: the emotional
    expression should match the regulation strategy decided in the empathy step,
    and the actual dialogue and actions should follow through on the objectives
    set in the construction step.  A thinking block that plans one strategy but
    an answer that executes another—or an emotional trajectory that shifts
    without justification between the two blocks—is a significant weakness.
\end{itemize}

\paragraph{Answer Rubric (B1--B4).}
The answer rubric evaluates the \texttt{<answer>} block along four dimensions (Table~\ref{tab:answer_rubric}):

\begin{itemize}[leftmargin=1.5em,itemsep=4pt,topsep=2pt]
    \item \textbf{B1. Conversational Ability.}  Evaluates fundamental
    linguistic quality: fluent and natural phrasing that reads like a human
    rather than a template, clear and easy-to-parse expression without awkward
    repetition or broken grammar, and logical continuity with the dialogue
    history.  Additionally, the bracketed structure must be correctly
    used—parentheses \texttt{(...)} for externally observable cues (actions,
    expressions, gestures) and brackets \texttt{[...]} for internal private
    content (thoughts, emotional reactions)—with untagged text reserved for
    spoken dialogue.  Inverting the bracket semantics is treated as a severe
    violation that caps the overall score.

    \item \textbf{B2. Character Consistency.}  Measures fidelity to the persona
    definition: identity and profile alignment (behavior and attitudes match
    the character's background and traits), knowledge accuracy (the character
    demonstrates only what it should know), voice and style fidelity (verbal
    habits, tone, and emotional baseline remain persona-consistent), and
    motivation coherence (stances and choices align with the character's drives
    and constraints).  Actions in \texttt{(...)} and inner descriptions in
    \texttt{[...]} must also reflect the character's unique personality rather
    than generic or out-of-character content.

    \item \textbf{B3. Interpersonal Interaction.}  Assesses social
    responsiveness: whether the reply directly addresses the user's message
    (both content and implied intent) rather than offering generic filler;
    whether tone and social distance match the defined relationship; whether
    the character adaptively acknowledges emotions, tension, or subtext; and
    whether the character refrains from speaking, acting, or generating
    thoughts on behalf of the user or any third-party NPC.  Subtle body
    language or inner reactions that show active engagement with the user's
    words are considered strengths.

    \item \textbf{B4. Narrative Progression.}  Evaluates whether the reply
    creates engaging forward momentum: adding meaningful progression (new
    actionable hooks, plot movement, or concrete next beats), demonstrating
    fitting emotional intelligence (empathy, tension control, appropriate
    reaction), and exhibiting expressive variety with vivid but controlled
    phrasing.  The reply should leave a clear conversational handle—a question,
    invitation, proposal, or pointed response—that invites continuation.
    Actions or internal realizations that actively advance the scene or shift
    the dynamic are considered strengths.
\end{itemize}


\section{CharacterBench Detailed Results}
\label{app:characterbench}

Table~\ref{tab:characterbench_detail} presents the full per-dimension
results on CharacterBench.  The six dimensions---Memory, Knowledge,
Person, Emotion, Morality, and Believability---are each averaged from
their respective sub-metrics as defined by the benchmark.

\begin{table}[htbp]

\caption{Per-dimension results on CharacterBench. Memory = Memory Consistency; Knowledge = avg.\ of Fact Accuracy \& Boundary Consistency; Person = avg.\ of Attribute Consistency (Bot/Human) \& Behavior Consistency (Bot/Human); Emotion = avg.\ of Emotional Self-regulation \& Empathetic Responsiveness; Morality = avg.\ of Morality Stability \& Morality Robustness; Believability = avg.\ of Human-likeness \& Engagement.}
\label{tab:characterbench_detail}

\centering
\footnotesize
\setlength{\tabcolsep}{3.5pt}
\renewcommand{\arraystretch}{1.05}
\begin{tabular}{lccccccc}
\toprule
\textbf{Model} & \textbf{Memory} & \textbf{Knowledge} & \textbf{Person} & \textbf{Emotion} & \textbf{Morality} & \textbf{Believability} & \textbf{Avg.} \\
\midrule
Doubao-1.5-Pro-Character & 3.83 & 3.41 & 4.06 & 3.18 & 4.88 & 3.42 & 3.84 \\
GPT-4o-mini & 3.72 & 3.18 & 4.24 & 3.23 & 4.90 & 3.45 & 3.86 \\
GPT-5-Chat & 3.52 & 3.22 & 4.21 & 3.21 & 4.91 & 3.63 & 3.87 \\
DeepSeek-V4-Flash & 3.18 & 3.27 & 3.87 & 2.99 & 4.83 & 3.31 & 3.65 \\
\midrule
Qwen3-4B-Instruct & 3.94 & 2.99 & 4.22 & 3.00 & 4.91 & 3.43 & 3.80 \\
\midrule
Qwen2.5-7B-Instruct & 3.64 & 2.95 & 3.94 & 3.12 & 4.87 & 3.03 & 3.64 \\
\quad + Psy-CoT + GRPO & 3.77 & 3.02 & 3.98 & 2.95 & 4.92 & 3.24 & 3.68 \\
\quad + Psy-CoT + GSPO & 3.77 & 2.98 & 3.91 & 2.93 & 4.90 & 3.26 & 3.65 \\
\quad + Psy-CoT + DAPO & 3.67 & 3.08 & 3.95 & 3.04 & 4.94 & 3.25 & 3.70 \\
\quad + Psy-CoT + RAPO & 3.77 & 3.03 & 4.07 & 3.07 & 4.92 & 3.34 & \textbf{3.75} \\
\midrule
Qwen2.5-7B-Instruct-CoSER & 4.02 & 2.94 & 3.67 & 2.86 & 4.69 & 2.95 & 3.50 \\
Qwen2.5-7B-Instruct-Crab & 4.16 & 2.78 & 3.85 & 3.12 & 4.80 & 3.07 & 3.62 \\
Qwen2.5-7B-Instruct-AdaMARP & 4.03 & 3.00 & 3.99 & 3.07 & 4.80 & 3.04 & 3.68 \\
\midrule
Llama-3.1-8B-Instruct & 3.65 & 3.16 & 4.12 & 3.25 & 4.84 & 3.36 & 3.79 \\
\midrule
Llama-3.1-8B-Crab & 4.25 & 2.80 & 3.90 & 3.14 & 4.79 & 3.07 & 3.65 \\
Llama-3.1-8B-CoSER & 4.15 & 2.92 & 3.92 & 3.00 & 4.88 & 3.00 & 3.64 \\
\midrule
Qwen2.5-14B-Instruct & 4.02 & 3.11 & 4.06 & 3.26 & 4.93 & 3.13 & 3.77 \\
Qwen2.5-14B-Instruct-AdaMARP & 4.04 & 3.02 & 3.83 & 3.04 & 4.80 & 3.17 & 3.64 \\
Llama-3.1-70B-CoSER & 3.85 & 3.18 & 3.92 & 3.17 & 4.81 & 3.19 & 3.71 \\
Llama-3.3-70B-Instruct & 3.71 & 3.38 & 4.18 & 3.23 & 4.91 & 3.39 & 3.85 \\
\bottomrule
\end{tabular}

\end{table}


\section{CharacterEval Detailed Results}
\label{app:charactereval_detail}

Table~\ref{tab:charactereval_detail} presents the per-dimension results on CharacterEval. Conversational Ability = avg.\ of Coherence, Consistency \& Fluency; Character Consistency = avg.\ of Exposure, Accuracy, Hallucination, Behavior \& Utterance; Role-Playing Attractiveness = avg.\ of Humanlikeness, Communication Skills, Diversity \& Empathy.

\begin{table}[htbp]

\caption{Per-dimension results on CharacterEval. Coh = Coherence; Con = Consistency; Flu = Fluency; Exp = Exposure; Acc = Accuracy; Hal = Hallucination; Beh = Behavior; Utt = Utterance; Hli = Humanlikeness; Com = Communication Skills; Div = Diversity; Emp = Empathy.}
\label{tab:charactereval_detail}

\centering
\footnotesize
\setlength{\tabcolsep}{2.5pt}
\begin{tabular}{lccccccccccccc}
\toprule
& \multicolumn{3}{c}{\textbf{Conv.\ Ability}} & \multicolumn{5}{c}{\textbf{Character Consistency}} & \multicolumn{4}{c}{\textbf{Role-Playing Attractiveness}} & \\
\cmidrule(lr){2-4} \cmidrule(lr){5-9} \cmidrule(lr){10-13}
\textbf{Model} & \textbf{Coh} & \textbf{Con} & \textbf{Flu} & \textbf{Exp} & \textbf{Acc} & \textbf{Hal} & \textbf{Beh} & \textbf{Utt} & \textbf{Hli} & \textbf{Com} & \textbf{Div} & \textbf{Emp} & \textbf{Avg} \\
\midrule
Doubao-1.5-Pro-Character & 4.14 & 3.97 & 3.77 & 2.51 & 3.19 & 3.25 & 3.79 & 3.33 & 3.73 & 3.64 & 3.23 & 3.42 & 3.50 \\
GPT-4o-mini & 3.93 & 3.50 & 3.52 & 2.66 & 2.99 & 3.10 & 3.34 & 3.07 & 3.12 & 3.68 & 2.65 & 3.31 & 3.23 \\
GPT-5-Chat & 4.08 & 3.99 & 3.83 & 2.50 & 3.15 & 3.23 & 4.05 & 3.35 & 3.88 & 3.25 & 3.56 & 3.21 & 3.50 \\
DeepSeek-V4-Flash & 4.09 & 3.86 & 3.73 & 2.62 & 3.14 & 3.24 & 3.78 & 3.26 & 3.68 & 3.75 & 3.13 & 3.40 & 3.47 \\
\midrule
Qwen2.5-7B-Instruct & 3.87 & 3.56 & 3.53 & 2.29 & 2.93 & 2.94 & 2.99 & 3.01 & 3.31 & 3.32 & 2.39 & 3.16 & 3.09 \\
\quad + Psy-CoT + GRPO & 3.33 & 3.20 & 3.62 & 3.04 & 3.14 & 3.37 & 2.74 & 3.70 & 3.06 & 3.81 & 3.43 & 3.09 & 3.29 \\
\quad + Psy-CoT + GSPO & 3.77 & 3.28 & 3.44 & 2.81 & 3.15 & 3.06 & 3.64 & 2.95 & 2.96 & 3.69 & 3.12 & 3.37 & 3.27 \\
\quad + Psy-CoT + DAPO & 3.80 & 3.44 & 3.47 & 2.62 & 3.14 & 3.01 & 3.70 & 3.05 & 3.19 & 3.50 & 3.06 & 3.32 & 3.27 \\
\quad + Psy-CoT + RAPO & 3.87 & 3.58 & 3.62 & 2.55 & 3.17 & 3.07 & 3.81 & 3.09 & 3.37 & 3.35 & 3.24 & 3.35 & \textbf{3.33} \\
\midrule
Qwen2.5-7B-Instruct-CoSER & 3.75 & 3.71 & 3.42 & 1.76 & 2.80 & 2.75 & 2.55 & 3.01 & 3.69 & 2.44 & 2.03 & 2.84 & 2.87 \\
Qwen2.5-7B-Instruct-Crab & 3.68 & 3.62 & 3.42 & 1.74 & 2.80 & 2.76 & 2.53 & 2.94 & 3.54 & 2.49 & 2.06 & 2.83 & 2.84 \\
Qwen2.5-7B-Instruct-AdaMARP & 3.74 & 3.62 & 3.42 & 1.75 & 2.81 & 2.76 & 2.52 & 2.94 & 3.57 & 2.50 & 2.02 & 2.81 & 2.84 \\
\midrule
Qwen3-4B-Instruct & 3.88 & 3.60 & 3.51 & 2.81 & 3.15 & 3.17 & 4.05 & 3.19 & 3.52 & 3.39 & 3.75 & 3.23 & 3.44 \\
\midrule
Llama-3.1-8B-Instruct & 3.66 & 3.41 & 3.31 & 2.08 & 2.74 & 2.76 & 3.12 & 2.86 & 3.17 & 2.90 & 2.58 & 2.87 & 2.94 \\
Llama-3.1-8B-Crab & 3.52 & 3.39 & 3.21 & 1.64 & 2.66 & 2.57 & 2.32 & 2.78 & 3.35 & 2.28 & 1.93 & 2.67 & 2.67 \\
Llama-3.1-8B-CoSER & 3.47 & 3.14 & 3.14 & 1.79 & 2.60 & 2.53 & 2.65 & 2.59 & 3.10 & 2.49 & 2.17 & 2.67 & 2.68 \\
\midrule
Qwen2.5-14B-Instruct & 3.89 & 3.58 & 3.56 & 2.35 & 3.02 & 3.00 & 3.09 & 3.06 & 3.44 & 3.20 & 2.47 & 3.18 & 3.13 \\
Qwen2.5-14B-Instruct-AdaMARP & 3.85 & 3.83 & 3.52 & 1.70 & 2.78 & 2.83 & 2.81 & 3.05 & 3.74 & 2.49 & 2.20 & 2.83 & 2.94 \\
\bottomrule
\end{tabular}

\end{table}


\section{Training Hyperparameters}
\label{app:hyperparams}

For reward evaluation, we deploy Qwen3-30B-A3B-Instruct-2507-FP8 on two high-memory GPUs to increase inference concurrency, with the judge concurrency set to 512. For policy training, we use six high-memory GPUs to train the role-playing model. In our setup, 100 training steps take approximately 24 hours.

Table~\ref{tab:hyperparams} summarizes the hyperparameters used in RAPO
training.

\begin{table}[h]

\caption{RAPO training hyperparameters.}
\label{tab:hyperparams}

\centering
\small
\begin{tabular}{ll}
\toprule
Hyperparameter & Value \\
\midrule
Learning rate & $1 \times 10^{-6}$ \\
Train batch size & 288 \\
PPO mini-batch size & 144 \\
Negative cap $\mu_{-}$ & 0.1 \\
\midrule
\multicolumn{2}{l}{\textbf{Reward}} \\
Answer length penalty zone & 300--350 chars (linear) \\
\bottomrule
\end{tabular}

\end{table}

\refstepcounter{table}\label{tab:think_rubric}
\begin{promptbox}[Table~\thetable: Think Rubric --- Full Evaluation Prompt (A1--A4)]
\begin{lstlisting}[style=promptstyle]
You are a strict evaluator of a role-playing model's <system_think> block (the model's internal reasoning before producing an in-character response).

Evaluate ONLY the <system_think> block. Do not evaluate the <answer>.

Generate one coherent weakness paragraph and one **scalar score** in [0, 1].

Conservative scoring policy:
- Do NOT award above 0.6 unless there is clear and compelling evidence across ALL evaluation criteria below
- 0.8+ should be rare
- 1.0 is reserved for thinking blocks that are essentially flawless

============================================================
A1. Interaction Perception
============================================================
Definition:
Evaluates the thinking block's comprehension of both the global context and the immediate conversational moment.
Checklist:
- Global view: demonstrates accurate understanding of the current physical and social setting - the environment, spatial relationships, who is present, and the broader scene dynamics at play
- Recent-turn perception: correctly identifies the user's most recent explicit content, primary intent, and any implicit cues (emotional tone, subtext, unspoken needs, pressure)
- Deep reading: goes beyond surface-level parsing to recognize the user's underlying psychological state, shifting dynamics, or hidden stakes in the latest turn
- Multi-turn continuity: tracks references to prior events, evolving emotional states, and relationship changes across the dialogue
- If the thinking misses or misreads the global setting or a critical cue from the most recent turn, this is a major weakness

============================================================
A2. Psychological Empathy
============================================================
Definition:
Assesses whether the thinking block faithfully models the character's internal emotional and psychological life based on their established persona - not a generic empathetic response, but one rooted in the specific character's background, values, flaws, and relational history.
Checklist:
- Persona alignment: the emotional reasoning is consistent with the character's unique background, core values, personality flaws, and established behavioral patterns - not a generic "understanding" that any character could produce
- Relational specificity: considers the character's past relationship and history with the current addressee, including any unresolved tensions, established bonds, power dynamics, or emotional baggage
- Cognitive blind spots: the thinking retains the character's own biases, prejudices, and blind spots - it does not achieve an unrealistic level of self-awareness that contradicts the persona. A flawed character should think like a flawed character, not like an omniscient therapist
- Coping mechanisms: the thinking reflects how much of the character's internal reaction should be openly expressed, deliberately masked, or even weaponized - based purely on the character's established coping patterns, not on what would be "healthiest" or "most reasonable"
- If empathy is generic, performative, or violates the character's established psychological profile, this is a significant weakness

============================================================
A3. Logical Construction
============================================================
Definition:
Evaluates the coherence and quality of the reasoning process - whether the thinking block synthesizes available information, applies the character's worldview, and produces action plans that are genuinely character-driven.
Checklist:
- Fact synthesis: integrates all relevant facts from the dialogue context, the scene, and the character's knowledge - important details are not overlooked or forgotten
- Worldview application: reasoning is filtered through the character's unique worldview, beliefs, and priorities rather than arriving at objectively "optimal" conclusions
- Goal authenticity: the short-term goals and planned actions are the closest match to what this specific character would actually do given their profile, motivations, and current emotional state - not what a perfectly rational or morally ideal character would do
- Coherent decision chain: thinking follows a logical path from perception through analysis to decision to planned response, without circular reasoning or unexplained leaps
- Constraint awareness: considers relevant limitations (character knowledge, physical constraints, social norms, scene conditions) in planning
- If reasoning is shallow, ignores key facts, or produces goals that feel generic or out-of-character, this is a significant weakness

============================================================
A4. Think-Answer Consistency
============================================================
Definition:
Evaluates whether the thinking block is internally coherent and consistent - whether the reasoning, emotional strategy, and planned actions form a logical and character-authentic chain.
Checklist:
- The thinking's internal logic should be consistent: perception leads to analysis, analysis leads to decisions, decisions lead to planned actions
- The emotional trajectory should be coherent - the character's feelings should evolve naturally
- If the thinking is internally contradictory, circular, or incoherent, this is a significant weakness

============================================================
Weakness Writing Rules
============================================================
- Write one coherent weakness paragraph evaluating the <system_think> block.
- The paragraph should flow naturally as continuous prose - do NOT use bullet points, numbered lists, or per-dimension sub-headings.
- Cover all evaluation criteria (A1-A4) in your paragraph. Weave them together logically, referencing specific evidence from the text.
- If there is no meaningful weakness, output an empty string "".

============================================================
Output Format (Strict JSON only)
============================================================
Return ONLY the following JSON object (no extra keys, no markdown):
{
  "weakness": "coherent paragraph evaluating think quality...",
  "score": <float 0-1>
}
\end{lstlisting}
\end{promptbox}

\refstepcounter{table}\label{tab:answer_rubric}
\begin{promptbox}[Table~\thetable: Answer Rubric --- Full Evaluation Prompt (B1--B4)]
\begin{lstlisting}[style=promptstyle]
You are a strict evaluator of a role-playing model's <answer> block (the in-character response).

Evaluate ONLY the <answer> block. Do not evaluate the <system_think>.

Generate one coherent weakness paragraph and one **scalar score** in [0, 1].

Conservative scoring policy:
- Do NOT award above 0.6 unless there is clear and compelling evidence across ALL evaluation criteria below
- 0.8+ should be rare
- 1.0 is reserved for answers that are essentially flawless

============================================================
B1. Conversational Ability
============================================================
Definition:
Evaluates fundamental linguistic quality and long-turn conversational stability: fluency, naturalness, and continuity with prior turns.
Checklist:
- Fluent and natural phrasing; reads like a human, not template-like or robotic
- Clear and easy to parse; avoids awkward repetition or broken grammar
- Maintains logical continuity with dialogue history; no internal inconsistencies
- Bracketed structure correctly used:
  - `()` for externally observable content - actions, expressions, posture, tone of voice, gestures, and any other behavior that others can see, hear, or perceive
  - `[]` for internally private content - inner thoughts, mental monologue, self-reflection, emotional reactions, and anything others cannot perceive
  - Untagged text is spoken dialogue
  - **SEVERE VIOLATIONS (hard score cap): putting externally observable content inside `[]` or internally private content inside `()` - this fundamentally inverts the bracket semantics. If this violation occurs, the overall output `score` MUST NOT exceed 0.2, regardless of performance on all other dimensions.**

============================================================
B2. Character Consistency
============================================================
Definition:
Measures fidelity to persona definition, including identity, knowledge level, style, and motivation coherence.
Checklist:
- Identity/profile fidelity: behavior and attitudes match character background and traits
- Knowledge accuracy: demonstrates only what the character should know
- Voice/style fidelity: verbal habits, tone, and emotional baseline remain persona-consistent
- Motivation coherence: stance and choices align with character drives and constraints
- Actions in `()` and psychological descriptions in `[]` must match the character's unique personality and behavioral patterns. Generic or out-of-character physical gestures and inner monologue are weaknesses.

============================================================
B3. Interpersonal Interaction
============================================================
Definition:
Assesses social responsiveness: how directly and appropriately the character engages with the user's content, intent, and relational context.
Checklist:
- Directly addresses user's message (content + implied intent), not generic filler
- Relationship awareness: tone and social distance match the defined relationship
- Social dynamics are adaptive (acknowledging emotions, tension, stakes, or subtext when present)
- Does not speak, act, or generate thoughts on behalf of the user or any third-party NPC
- If actions in `()` or psychological descriptions in `[]` enhance the sense of interaction and engagement (e.g., subtle body language responding to the user's words, inner reactions that show active listening), this is a strength that can raise the score.

============================================================
B4. Narrative Progression
============================================================
Definition:
Evaluates whether the reply creates engaging forward momentum: emotional intelligence, expressive variety, and hooks that invite continuation.
Checklist:
- Adds meaningful progression (new actionable hooks, plot movement, or concrete next beats)
- Demonstrates fitting emotional intelligence (empathy, tension control, appropriate reaction)
- Expression diversity: vivid but controlled phrasing without verbosity
- Leaves a clear next conversational handle (question, invitation, proposal, or pointed response)
- If actions in `()` or psychological descriptions in `[]` actively advance the scene or plot (e.g., a decisive gesture that shifts the dynamic, an internal realization that changes the character's stance), this is a strength that can raise the score.

============================================================
Weakness Writing Rules
============================================================
- Write one coherent weakness paragraph evaluating the <answer> block.
- The paragraph should flow naturally as continuous prose - do NOT use bullet points, numbered lists, or per-dimension sub-headings.
- Cover all evaluation criteria (B1-B4) in your paragraph. Weave them together logically, referencing specific evidence from the text.
- If there is no meaningful weakness, output an empty string "".

============================================================
Output Format (Strict JSON only)
============================================================
Return ONLY the following JSON object (no extra keys, no markdown):
{
  "weakness": "coherent paragraph evaluating answer quality...",
  "score": <float 0-1>
}
\end{lstlisting}
\end{promptbox}


\section{New Assets and Safeguards for Model Release}
\label{app:new_assets_safeguards}

We do not introduce any new dataset in this work. Our new asset is a role-playing model trained by fine-tuning existing open-source base models, and we plan to release this model after acceptance. Since this manuscript itself serves as the technical report, the training setup and major hyperparameters are documented in Appendix~\ref{app:hyperparams}, and the known limitations are discussed in Section~\ref{sec:ablation}.

For responsible release, we will publish explicit usage constraints and clarify that misuse-oriented scenarios are outside the intended scope. We will adopt the MIT license for the released model artifacts and provide transparent release notes on intended use and limitations to support reproducibility and risk-aware adoption.

\section{Human Subjects and Ethics Statement}
\label{app:human_subjects_ethics}

This human evaluation (Section~\ref{sec:human_evaluation}) involved minimal risk. Participants were informed of potential risks and provided consent before annotation; they could withdraw at any time. No sensitive personal data were collected. The study complied with the institution's human-subject research requirements (IRB/equivalent approval or exemption) prior to data collection.

\section{Broader Impacts}
\label{app:broader_impacts}

Our work has both potential positive and negative societal impacts: on the positive side, we identify several general deficiencies in reinforcement learning training for role-playing tasks, and we propose a stronger psychology-grounded chain-of-thought framework (Psy-CoT) and a role-aware policy optimization algorithm (RAPO), which extend existing role-playing methodologies and provide practical solutions to previously under-addressed issues in character fidelity, interaction quality, and training effectiveness; on the negative side, improved role-playing capability may be misused in non-compliant scenarios (e.g., deceptive or manipulative role simulation), so we encourage responsible use, compliance with platform and legal policies, and risk-aware deployment practices.

\end{document}